%% file: paper_pre.tex
\pdfoutput=1
\documentclass[]{revtex4-1}

\usepackage{graphicx}
\usepackage{amsmath}
\usepackage{bm}
\usepackage{algcompatible}

\usepackage{algpseudocode}

\usepackage{subcaption}

\usepackage{url}

\usepackage{newfloat}
\usepackage[size=small]{caption}
\usepackage{etoolbox}

\makeatletter
  \def\vhrulefill#1{\leavevmode\leaders\hrule\@height#1\hfill \kern\z@}
\makeatother

\AtEndEnvironment{algorithm}{\vskip-8pt\noindent\vhrulefill{0.7pt}\par\nobreak\vskip2pt}

\DeclareFloatingEnvironment[
    fileext=loa,
    listname=List of Algorithms,
    name=ALGORITHM,
    placement=tbhp,
]{algorithm}
\DeclareCaptionFormat{algorithms}{\vskip-10pt\vhrulefill{0.7pt}\vskip-1.5pt\par#1#2#3\vskip-6.5pt\hrulefill\vskip-8pt}
\algblock[Input]{Input}{EndInput}
\algblockdefx[Input]{Input}{EndInput}%
    [1]{\textbf{Input} #1}%
    {}

\newcommand{\fig}{\figurename~}
\newcommand{\tab}{\tablename~}
\newcommand{\alg}{ALGORITHM~}

\newcommand{\shortcite}{\cite}

\newcommand{\seqsize}{0.25}

\newcommand{\return}{ }

\begin{document}

\input{paper_abst.tex}

\title{Stitching Stabilizer:\\Two-frame-stitching Video Stabilization for Embedded Systems}
\author{Masaki Satoh}
\email{m-satoh@morphoinc.com}
\affiliation{Morpho, inc.}
\date{\today}

\maketitle

\newif\ifTEASER
\TEASERfalse

\input{paper_impl.tex}


\bibliography{paper}

\end{document}

%% file: paper_abst.tex
\begin{abstract}
 In conventional electronic video stabilization, the stabilized
 frame is obtained by cropping the input frame to cancel camera shake.
 While a small cropping size results in strong stabilization,
 it does not provide us satisfactory results
 from the viewpoint of image quality, because 
 it narrows the angle of view.
 By fusing several frames, we can effectively expand the area of input frames,
 and achieve strong stabilization even with a large cropping size.
 Several methods for doing so have been studied.
 However, their computational costs
 are too high for embedded
 systems such as smartphones.

 We propose a simple, yet surprisingly effective algorithm, called the stitching
 stabilizer.
 It stitches only two frames
 together
 with a minimal computational cost.
 It can achieve real-time processes in embedded systems, for
 Full~HD and 30 FPS videos.
 To clearly show the effect,
 we apply it to
 hyperlapse.
 Using several clips, we show it produces more strongly
 stabilized and natural results than the existing solutions from Microsoft and Instagram.


\end{abstract}

%% file: paper_impl.tex
\section{Introduction}

Smartphones have recently become very popular worldwide.
For consumers, one of the criteria for choosing a handset
is the camera features.
Although the capability of taking beautiful photos is an important factor, 
the 
video quality should not be overlooked.
%
There are many features that affect the overall video quality of smartphones 
such as auto focus, auto exposure and auto white balance.
One important feature is stabilization.

Three are major video stabilization mechanisms, namely
mechanical image stabilization, optical image stabilization~(OIS), and
electronic image stabilization~(EIS).
EIS is a practical solution for embedded systems such as smartphones, 
because it is inexpensive and requires no special
hardware such as gyroscope sensors or optical systems.
The simplest algorithm for EIS 
creates output frames by cropping the input in a
way that cancels camera shake.
One advantage of this algorithm is its simplicity.
Only limited resources are available for embedded systems,
so simplicity is crucial to achieve real-time processes such as
video stabilization.
Moreover, this is a GPU friendly algorithm, which has become an important
factor in recent technology trends.
Cropping, or geometrical transformation, of image frames is a strong
point of the GPU.

EIS does have a weak point, however.
That is, it sacrifices the angle of view~(AOV) of
the camera, which is important for quality.
This is an inevitable consequence of cropping.
Less cropping is advantageous for stabilization but disadvantageous for the AOV.
Hence, we must tune and find the best balance for the cropping size.
%

Let us consider an EIS system with a large cropping size.
Although the AOV is wide in that case, there is no room to absorb camera
shake.
If we want stabilized results even in this situation, it is necessary to
accept cropping outside the input frame.
As there are no defined pixels outside the frame, the result must contain
undefined or deficit regions as depicted in \fig \ref{teaser}~(a).
If there was a method to fill in this deficit, we could obtain the stabilized
result with a wide AOV.
The filling process corresponds to effectively expanding the area of input frames, and
several techniques have been developed that try to achieve this.
%

\ifTEASER
\else
\begin{figure}[htbp]
   \centering
  \begin{subfigure}[t]{.317 \hsize}
  \includegraphics[scale=0.25]{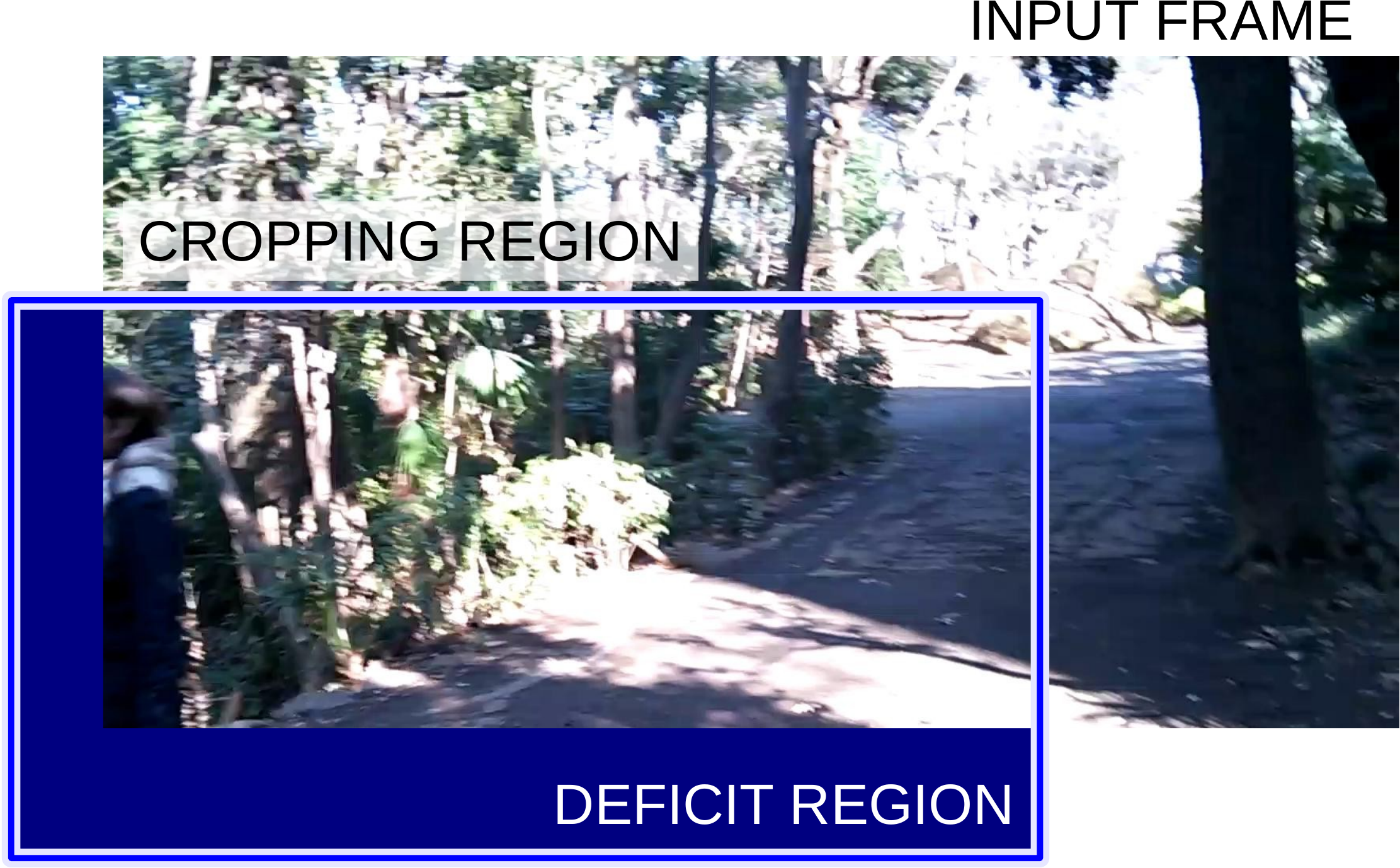}
   \caption*{(a) No-stitching stabilizer.}
  \end{subfigure}
  \begin{subfigure}[t]{.335 \hsize}
  \includegraphics[scale=0.25]{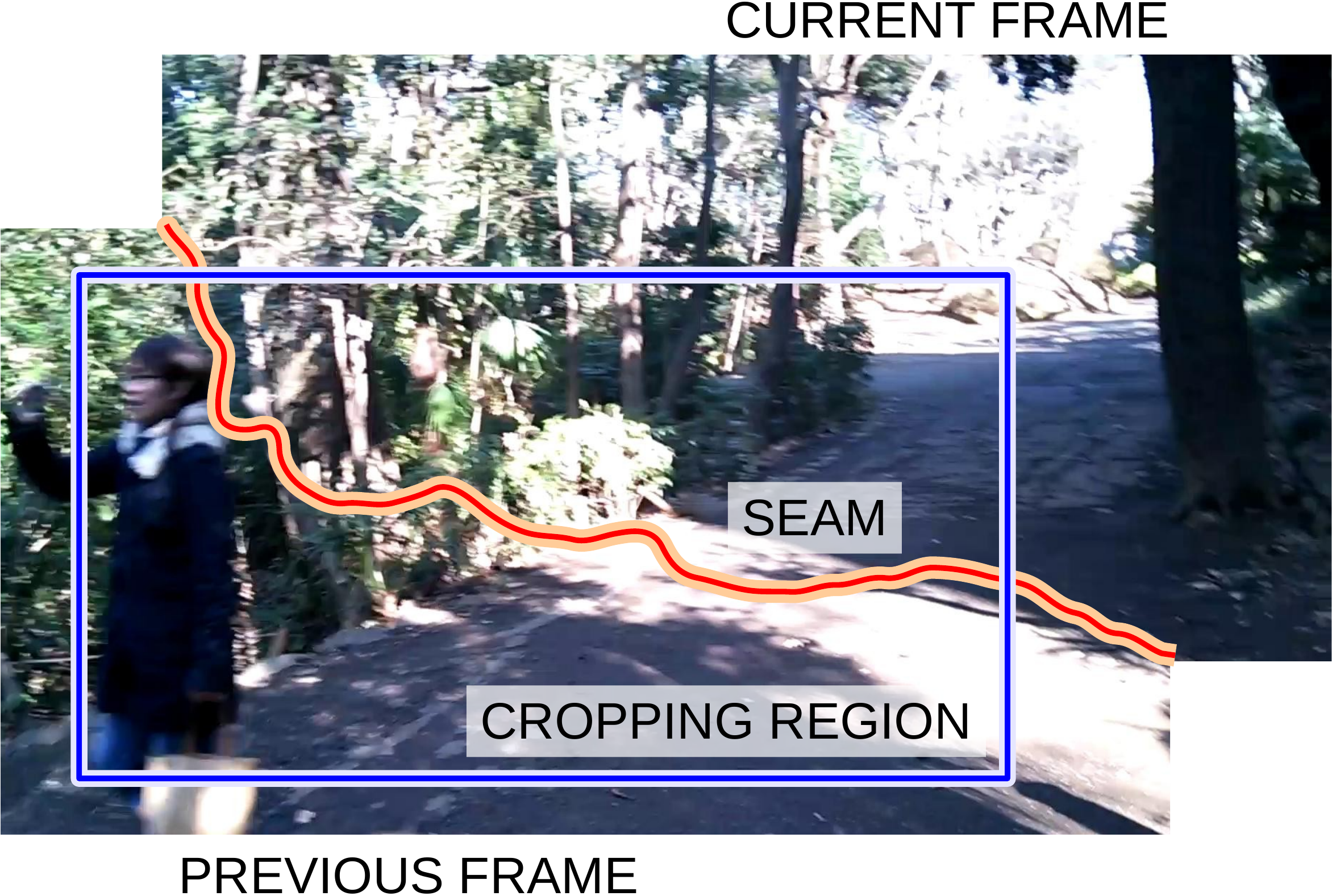}
   \caption*{(b) Stitching stabilizer.}
  \end{subfigure}
  \begin{subfigure}[t]{.23 \hsize}
  \includegraphics[trim={1 14cm 11cm 6cm},clip,scale=0.17]{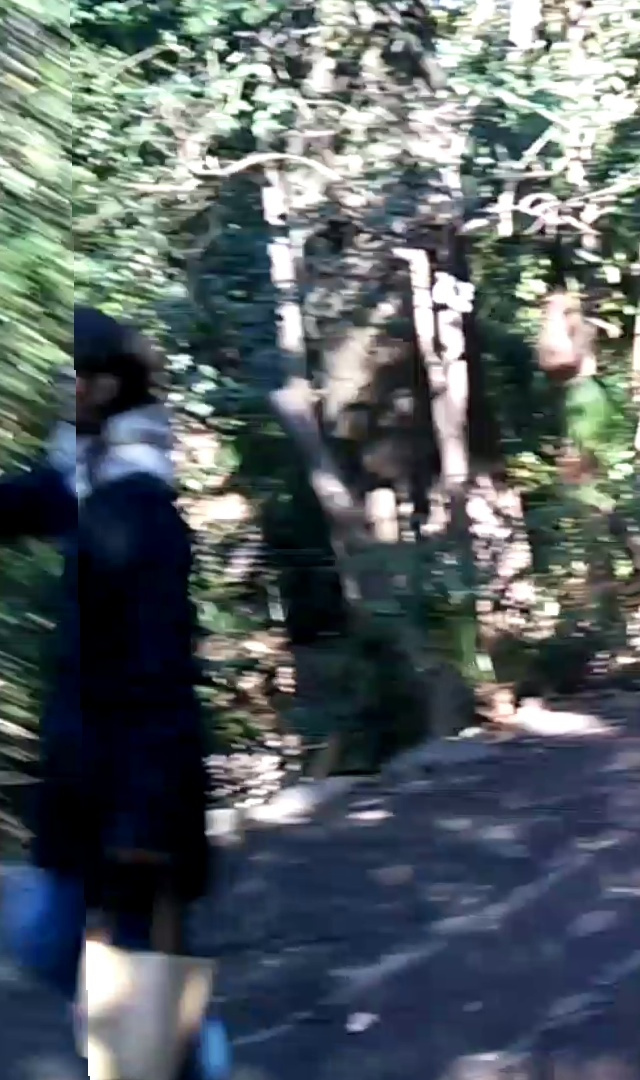}
  \includegraphics[trim={1 14cm 11cm 6cm},clip,scale=0.17]{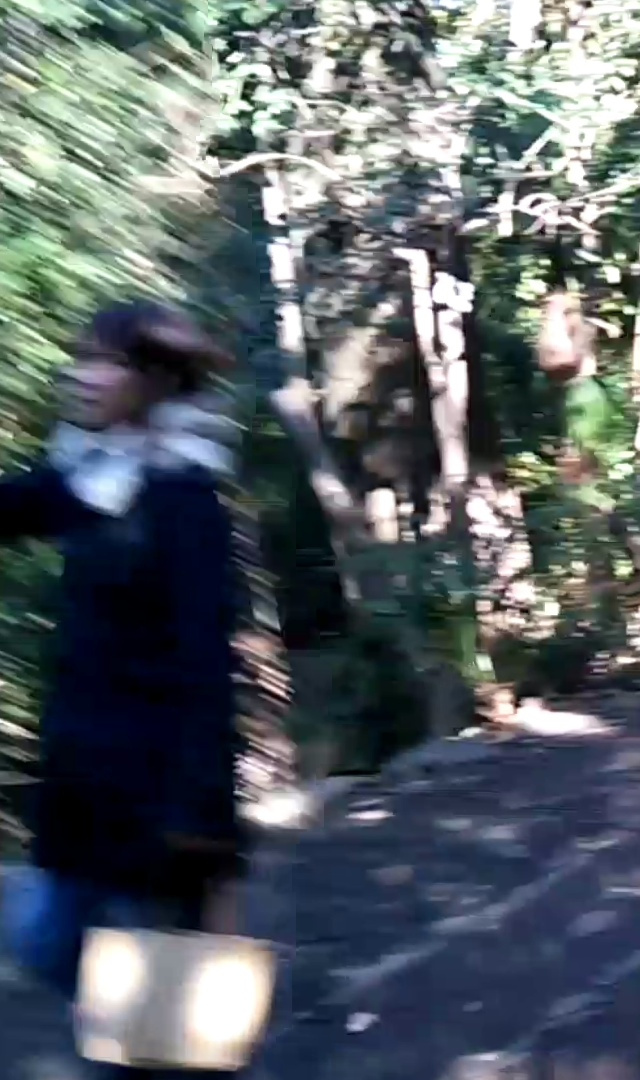}
   \caption*{(c) Stitching example.}
  \end{subfigure}
  \caption{
  (a) When we try to stabilize a large degree of camera shake with a large cropping
  size,
  there is a deficit region.
  This is because there are no defined pixels outside the frame.
  To achieve strong stabilization with a large cropping size, or a wide
  angle of view, we need to find some ways to
  fill in this deficit.
  (b) In our approach, the stitching stabilizer, the current and previous frames are stitched together along the
  seam to effectively expands the area of input frames.
  This fills the deficit in and strong stabilization with a wide angle
  of view is achieved.
  The algorithm searches for the seam so that stitching becomes as unnoticeable as
  possible.
  In this example, the seam should avoid the face of a person.
  (c) In the left panel,
  the undefined region is simply filled by the previous frame and the
  face is cut in half.
  The
  right panel is our stitching result.
  The seam avoids the face and the result is more natural.
  }
  \label{teaser}
\end{figure}
\fi

Litvin et al.~\shortcite{litvin2003probabilistic} employed mosaicing
from several adjacent frames to fill in the
deficit. The filling process is done by averaging corresponding valid
pixels. 
Their method does not consider non-planar or moving objects,
though, so
Matsushita et al.~\shortcite{matsushita2006full} 
and Chen et al.~\shortcite{chen2008capturing}
proposed methods that compensate for local motion in addition to
global or planar motion.
All three algorithms require several (more than two) input frames to
generate one output and perform complicated calculations for every
frame.
Hence, they are not applicable for embedded system because of the high computational cost.

Although the goal is different, the hyperlapse algorithm proposed by 
Kopf et al.~\shortcite{kopf2014first} shows a similar effect.
Hyperlapse is a kind of time-lapse video that is captured during
motion such as while walking or driving a car.
This has become a trend for mobile video, and some
apps~(\url{http://research.microsoft.com/en-us/um/redmond/projects/hyperlapseapps/}; 
\url{https://hyperlapse.instagram.com/}) for it have been
released in recent years.
They reconstruct a 3D world from video input and
virtually re-capture hyperlapse videos in their 3D world.
Although this solves the AOV problem and the results are impressive, 
it also has a high computational cost and thus cannot be applied to embedded systems.

\subsection{Our approach}
In this approach, we basically follow the same idea, namely filling
in the deficit, with an algorithm that is fast enough for embedded systems.
To achieve the fast process, we use only two frames for filling.
Under this limited condition, we cannot use methods applied in previous
studies~\cite{litvin2003probabilistic,matsushita2006full,chen2008capturing}.
Therefore, we take a new approach, {\it stitching stabilizer}, in which the two frames are simply
stitched together along an unnoticeable seam.
This is shown in \fig \ref{teaser}~(b).
%
The stitching process effectively expands the area of input frames,
and the deficit regions can be filled in.
This paper shows that stitching only two frames with a simple algorithm
can improve stabilization quality remarkably.
%

There are two ways to stitch.
One is to stitch the current and previous frames, and the other is to stitch
the current and the next~(subsequent) frames.
Both possibilities are considered in
the following discussions.
Then, the $(n+1)$-th frame is required to make the $n$-th result and
this introduces a latency of one frame.
This is a drawback, although it is outweighed by the positive effect.


To clearly show its effect, we applied the stitching stabilizer to hyperlapse videos.
In hyperlapse, camera shake is effectively increased because of
fast-forwarding.
Therefore, in the conventional stabilizer, a small
cropping region, or a narrow AOV, is unavoidable to achieve strong stabilization.
Here, {\it conventional} refers to the stabilizer without stitching.
We show that the stitching stabilizer
remarkably
reduced camera shake
compared to the conventional stabilizer, without losing the AOV.
Moreover, it produced a more stabilized and natural output than
existing real-time hyperlapse methods~\cite{karpenko2014introduction,joshi2015real}.



This paper is organized as follows.
Related work is summarized in section \ref{related_work}.
In section \ref{motion_detection_and_filtering}, we explain
the algorithm of the video stabilizer without stitching, and
the extension of it to the stitching stabilizer is given
in section \ref{stitching_stabilizer}.
We apply the stitching stabilizer to hyperlapse in section
\ref{hyperlapse}, and the 
results of applying our algorithm are explained and discussed in section \ref{results}.
Section \ref{conclusion} is the conclusion.

\section{Related work}
\label{related_work}
In this section, we summarize related work on video stabilization,
hyperlapse, video completion and image stitching.

\subsection{Video stabilization}
There are 2D and 3D models of video stabilization algorithms.
Although the structure-from-motion based 3D model is more 
powerful~\cite{liu2009content},
it consumes large computational resources and is not robust.
The 2D
model~\cite{litvin2003probabilistic,matsushita2006full,grundmann2011auto}
uses 
2D geometric transformation, such as affine or
perspective transformation, to describe its global motion.
Hence, it is fast and robust because of its simplicity and is popular in embedded systems.
Some methods lie in between the 2D and 3D models
\cite{liu2011subspace,goldstein2012video,liu2013bundled}.
In addition,
an algorithm that utilizes a hardware gyroscope has also been
studied~\cite{karpenko2011digital}.

Another important factor affecting video stabilization is rolling
shutter~(RS) distortion.
This distortion is related to the physical motion of the camera.
Hence, there are many algorithms for removing RS distortion by
modeling motion: translational motion along $x$ and $y$ axes~\cite{im2006digital},
affine transformation~\cite{cho2007affine}, row-by-row
motion~\cite{liang2008analysis,baker2010removing}, and 3D motion~\cite{forssen2010rectifying}.
These motion based methods require hardware-dependent parameters
to remove distortion, but algorithms that do not require prior knowledge
have also been developed~\cite{grundmann2012calibration,liu2013bundled}.

We prefer robustness as a solution for embedded systems.
For many casual users, there are no retakes.
It is therefore much more important to produce a robust result with
acceptable quality
than a perfect but not very robust one.
Therefore, we use 2D stabilization with perspective transformation.
Of course, affine transformation is more robust, but its quality is unsatisfactory.
For the same reason, we use the translational model for RS
distortion~\cite{im2006digital}.
In this case, the distortion can be written as a matrix and is easy to implement.

\subsection{Hyperlapse}
For hyperlapse, as stated,
the algorithm of Kopf et al.~\shortcite{kopf2014first}
achieves impressive results, but its computational cost is too high for our 
purposes.
Real-time algorithms have been studied by
Karpenko~\shortcite{karpenko2014introduction} for Instagram's app
and Joshi
et al.~\shortcite{joshi2015real} for Microsoft's mobile app.

Karpenko's approach is fairly
straightforward.
The algorithm combines the hardware gyroscope and image frames to calculate 2D global
motion
and stabilizes it.
To create hyperlapse, the input frame is
skipped according to a user set parameter, for example $\times 4$ or $\times 8$.
The algorithm by Joshi et al. is a bit tricky.
The global motion detection is purely software-based conventional
2D detection.
However, they select the input frames to achieve smoother results.
In other words, they do
not skip frames with even spacing.
By selecting frames, they can avoid very shaky frames and achieve strong
stabilization.
Poleg et al. created another frame selection algorithm~\cite{poleg2015egosampling}.
However, its computational cost is also too high to be implemented in embedded systems.

Our approach is between 
Kopf et al.
and
Karpenko.
%
It is fast and based on the conventional framework, while it utilizes several (in our
case two) input frames to produce output.
Although the frame selection is compatible with the stitching stabilizer,
we employ frame skipping with even spacing, as a start.

\subsection{Video completion and image stitching}
We have investigated various methods for filling in deficit regions.
One such method is video completion~\cite{wexler2004space,jia2004video,patwardhan2007video},
which is a technique to fill in missing parts of videos.
It has already been studied in the context of video stabilization~\cite{litvin2003probabilistic,matsushita2006full,chen2008capturing}.
The problem with these methods is that they are computationally expensive for our purpose.
Therefore, we used the second candidate, simple stitching of two adjacent frames, in
the stitching stabilizer.

Image stitching is a very common process in computer 
vision~\cite{davis1998mosaics,efros2001image,kwatra2003graphcut,gu2009new}.
The problem we have been focusing on  is finding the optimal seam.
In that sense, seam carving~\cite{avidan2007seam} and digital
photomontage~\cite{agarwala2004interactive} are somewhat related.

There are several popular seam finding algorithms, including
Dijkstra's algorithm~\cite{dijkstra1959note,davis1998mosaics,avidan2007seam},
dynamic programming~(DP)~\cite{efros2001image,gu2009new}, and
graph cuts~\cite{kwatra2003graphcut,agarwala2004interactive}.

We used Dijkstra's algorithm for the stitching stabilizer, and
the discussion of this is in subsection \ref{stitching_path}.

\section{Conventional video stabilizer}
\label{motion_detection_and_filtering}
We explain our algorithm of the conventional stabilizer.
We consider stabilization
with 2D perspective transformation.

As a motion detection algorithm, we use the pyramid image approach~\cite{bergen1992hierarchical}.
The process at each layer is a little different from the standard one.
First, we track sparsely selected 
Harris-Stephens feature
points~\cite{harris1988combined} independently
by block matching which minimizes the sum of absolute difference.
The initial vectors for each block matching process are determined by the transformation of the 
lower layer.
Second, we calculate the perspective transformation of this layer by 
the least squares method.

Our algorithm for EIS is indicated as \alg \ref{realistic_eis}.
\textproc{CalcMotion} is a function that calculates the perspective
transformation $M$ between two adjacent frames:
\begin{align}
 (x_\text{curr},y_\text{curr},1)^{\rm T}
 \sim
 M
 (x_\text{prev},y_\text{prev},1)^{\rm T},
\end{align}
where $(x_\text{curr},y_\text{curr},1)$ and
$(x_\text{prev},y_\text{prev},1)$ are corresponding coordinates in the current and
previous frames.
Based on this $M$ and the previous results,
\textproc{Filter} determines the
perspective matrix $P$ for cropping.
\textproc{Crop} is a function to perform cropping based on $P$.
\textproc{EnsureInside} requires more explanation than the other functions.
In our framework,
\textproc{Filter} itself does not ensure that the cropping region is inside the
frame. Hence, we need another function, in this case
\textproc{EnsureInside}, 
to do this.
\textproc{CalcCroppingBoundary} calculates the boundary region of the cropping.
%
While the cropping region is not inside the frame,
the matrix $P$ is repeatedly blended with the identity matrix.
This blending corresponds to bringing the
matrix to near the identity.
The constant $\varepsilon$ is the blending coefficient and we set
%
%
$\varepsilon = 0.01$.
\textproc{EnsureInside} adds unfiltered motion to the output.
Hence, when \textproc{IsInside} frequently returns false, the output
video becomes shaky.
\begin{algorithm}[htbp]
 \caption{Algorithm for EIS.}
 \label{realistic_eis}
 \begin{algorithmic}[1]
  \State $P \gets identity$
  \While{there is a frame to be processed}
  \State Load the next input frame
  \State $M \gets$ \Call{CalcMotion}{from previous to current}
  \State $P \gets$ \Call{Filter}{$M,P$}
  \State $P \gets$ \Call{EnsureInside}{$P$}
  \State \Call{Crop}{$P$}
  \EndWhile
  \Statex

  \Function{EnsureInside}{$P$}
  \State $crop \gets$ \Call{CalcCroppingBoundary}{$P$}
  \While{\Call{IsInside}{$crop$} is false}
  \State $P \gets $\Call{ToIdentity}{$P,\varepsilon$}
  \State $crop \gets$ \Call{CalcCroppingBoundary}{$P$}
  \EndWhile
  \State \Return $P$
  \EndFunction
  \Statex

  \Function{ToIdentity}{$P,\varepsilon$}
  \State $I \gets identity$
  \State Return $\varepsilon I + (1-\varepsilon) P$
  \EndFunction
  \Statex

  \Function{IsInside}{$crop$}
  \If{$crop$ is inside the input frame}
  \State \Return true
  \Else
  \State \Return false
  \EndIf
  \EndFunction
 \end{algorithmic}
\end{algorithm}

In this stabilization framework, the most important factor is how the function
\textproc{Filter} works.

\subsection{Rolling shutter distortion}

Before explaining the \textproc{Filter} algorithm, we summarize how
we treat RS distortion.
In a digital camera, RS distortion is a problem unique to CMOS sensors
without a mechanical shutter,
and CCD sensors are RS distortion-free.
Because CMOS sensors have recently become very popular in smartphones and
digital still cameras, this distortion needs to be addressed.
Moreover, the algorithm developed for CMOS sensors is also effective
 for CCD sensors,
 but not {\it vice versa}.

Let us define the coordinates on the distorted input frame as
$(x,y,1)$ and that on the ideal distortion-free frame as $(X,Y,1)$.
There must be a relationship between them, such as 
$(x,y,1)^{\rm T} = {\cal D}((X,Y,1)^{\rm T})$.
In general, this function ${\cal D}$ cannot be written in a matrix 
form.
However, when the global motion between adjacent frames 
is just a translation along the $x$ and $y$ axes with constant velocity~\cite{im2006digital}:
\begin{align}
 M =
 \begin{pmatrix}
  1 & 0 & t_x \\
  0 & 1 & t_y \\
  0 & 0 & 1
 \end{pmatrix}, 
\end{align}
${\cal D}$ can be written as a matrix:
\begin{align}
 D =
 \begin{pmatrix}
  1 & - t_x / H& 0 \\
  0 & 1 - t_y / H & 0 \\
  0 & 0 & 1
 \end{pmatrix} ^{-1},
\end{align}
where $H$ is an effective height for the sensor and
is equal to or larger than the actual height.
$H$ takes a large value for a fast scanning sensor and
a small value for a slow sensor.
In realistic scenes, $M$ is not a purely translational matrix.
It is composed of eight variables:
\begin{align}
 M =
 \begin{pmatrix}
  a & b & c \\
  d & e & f \\
  g & h & 1
 \end{pmatrix}, 
 \label{def_m}
\end{align}
where we set the right-bottom component to one, because the overall
factor is meaningless.
For the video, the most dominant part of the matrix $M$
might be the translations $c$ and $f$. 
Therefore, the following approximation seems to be a practical one:
\begin{align}
 D =
 \begin{pmatrix}
  1 & - c / H& 0 \\
  0 & 1 - f / H & 0 \\
  0 & 0 & 1
 \end{pmatrix} ^{-1},
 \label{eq_D}
\end{align}
where we assume that RS distortion is caused only by the
translational components of $M$.
Using this matrix, we can obtain the correspondence between the distorted and
distortion-free coordinates: $(x,y,1)^{\rm T} \sim D(X,Y,1)^{\rm T}$.
Note that the symbol ``$\sim$'' denotes that the right-hand side and the
left-hand side are equal up to an overall factor.

\subsection{Filter design}
\label{filter_design}
Here, we show how the function \textproc{Filter} works in our stabilizer.
Let us define the matrix $M_n$ that relates the
$(n-1)$-th input frame coordinates $(x_{n-1},y_{n-1},1)$ 
and the $n$-th input frame coordinates $(x_{n},y_{n},1)$:
\begin{align}
 (x_{n},y_{n},1)^{\rm T}
 &\sim
 M_{n}
 (x_{n-1},y_{n-1},1)^{\rm T}
 \nonumber \\
 &\sim
 D_n N_{n} D_{n-1}^{-1}
 (x_{n-1},y_{n-1},1)^{\rm T}.
 \label{drel_n}
\end{align}
This can be obtained by the function \textproc{CalcMotion} in \alg \ref{realistic_eis}.
We separate the RS distortion matrix $D_n$ and $D_{n-1}$ and get the
distortion-free matrix $N_{n}\equiv D_{n}^{-1} M_n D_{n-1}$.
As we have discussed, our expression for $D_n$, equation (\ref{eq_D}),
is an approximated one,
so $N_n$ is not exactly
distortion-free.
Even so,
it is still more manageable than $M_n$.

To process the $n$-th frame, the result of the $(n-1)$-th
frame is already known.
It is
the 2D perspective matrix $P_{n-1}$ which represents the relation
between the coordinates on 
the $(n-1)$-th output frame $(X_{n-1},Y_{n-1},1)$ and those on the $(n-1)$-th input frame $(x_{n-1},y_{n-1},1)$:
\begin{align}
 (x_{n-1},y_{n-1},1)^{\rm T}
 &\sim
 P_{n-1}
 (X_{n-1},Y_{n-1},1)^{\rm T}
 \nonumber \\
 &\sim
 D_{n-1}Q_{n-1}
 (X_{n-1},Y_{n-1},1)^{\rm T}.
 \label{rel_n-1}
\end{align}
This $P_{n-1}$ determines how the $(n-1)$-th input frame is cropped to
create the corresponding output frame. 
The
distortion-free cropping matrix $Q_{n-1}$ is defined as
$Q_{n-1}\equiv D_{n-1}^{-1} P_{n-1}$.

The filtering algorithm determines $Q_n$ based on
$N_0...N_n$ and $Q_0...Q_{n-1}$.
Then, because $D_n$ is obtained from $M_n$, we can get $P_n\sim D_nQ_n$ and
perform cropping.
We consider two extreme cases for filtering.
One is $Q_n\sim N_n Q_{n-1}$. From the above equations, we get:
\begin{align}
 (x_{n-1},y_{n-1},1)^{\rm T}
 \sim
 D_{n-1} Q_{n-1}
 (X_{n},Y_{n},1)^{\rm T}.
\end{align}
This is the relation between the $n$-th output coordinates and
$(n-1)$-th input coordinates.
This equation and equation (\ref{rel_n-1}) indicate that the output
frames are identical for the $(n-1)$-th and $n$-th frames.
Therefore, the result must be a video with no motion.
This interpretation is based on the unrealistic assumption that all
motions are written by the 2D perspective transformation.
Even so, it might be true even for actual situations that the stabilizer
removes almost every camera motion, including panning or tilting.
%
Another case is $Q_n\sim Q_{n-1}$.
In this case, the cropping matrix of the $n$-th frame is the same as
that of the $(n-1)$-th frame, except for the RS distortion.
This means there is no stabilization.
Needless to say, both cases are undesirable.
What we expect from video stabilization is stabilization for high
frequency motion and no stabilization for low frequency motion.
In other words, $Q_n\sim N_n Q_{n-1}$ for high frequency and $Q_n\sim Q_{n-1}$ for
low frequency.
Formally, we can write this as:
\begin{align}
 Q_n \sim f(N_n^{-1}) N_n Q_{n-1},
 \label{lpf}
\end{align}
where the function $f$ is a low pass filter.
This $f(N_n^{-1})$ represents how the camera moves in the output video, 
because $Q_n \sim N_n Q_{n-1}$ corresponds to the case
without camera motion.

To design the filter for $N_n^{-1}$, we first consider the most important
elements.
In video stabilization, these are the rotational angles.
While we follow the conventional yaw-pitch-roll notation,
our method is applicable for any of the Euler angles.
It is difficult to deduct exact angles from $N_n^{-1}$, because it is not
exactly distortion-free.
Hence, we simply estimate them.
For example, even the following rough estimation works:
\begin{align}
 \alpha_n = \sin^{-1}\frac{c'_n}{\ell},\quad
 \beta_n = -\sin^{-1}\frac{f'_n}{\ell},\quad
 \gamma_n = \tan^{-1}\frac{d'_n}{e'_n},
\end{align}
where $\alpha_n,\beta_n$, and $\gamma_n$ are the estimated yaw, pitch, and roll angles of
$N_n^{-1}$, respectively
and 
$\ell$ denotes the focal length of the camera.
The variables $c'_n,f'_n,d'_n$ and $e'_n$ are the components of the matrix
$N_n^{-1}$:
\begin{align}
 N_n^{-1} =
 \begin{pmatrix}
  a'_n & b'_n & c'_n \\
  d'_n & e'_n & f'_n \\
  g'_n & h'_n & 1
 \end{pmatrix}.
\end{align}
Then, we decompose $N_{n}^{-1}$ as follows:
\begin{align}
 N_{n}^{-1} &\sim R_{\rm yaw}(\alpha_{n})R_{\rm pitch}(\beta_{n}) 
 R_{\rm roll}(\gamma_{n}) L_{n},
 \label{decomposition}
\end{align}
where $R_{\rm yaw}(\alpha_n),R_{\rm pitch}(\beta_n)$, and 
$R_{\rm roll}(\gamma_n)$ are the 
matrix representation of the estimated yaw, pitch, and roll rotations.
The term $L_n$ includes all other motions and
errors of estimation.
In normal cases, this $L_n$ is a minor contributor to quality.
Therefore, we omit this term in the low pass filter:
\begin{align}
 f(N_n^{-1}) \sim R_{\rm yaw}(g(\alpha_{n}))R_{\rm pitch}(g(\beta_{n})) 
 R_{\rm roll}(g(\gamma_{n})),
 \label{filter_g}
\end{align}
where $g$ is a low pass filter for scalar values.
Again, $g(\alpha_n),g(\beta_n)$, and $g(\gamma_n)$ represent how the
camera moves.

For simplicity, we use the same type of filters for $\alpha_n,\beta_n$, and $\gamma_n$.
There are many possibilities for this filter algorithm.
Averaging recent values is a simple example.
Instead, we use the mid-range value that is defined as:
\begin{align}
 \text{mid-range value}(array) = \frac{\text{max}(array) + \text{min}(array)}{2},
\end{align}
where the variable $array$ is an array that stores recent input values.
The advantage of the mid-range value in comparison to the average value
is quick response, which is an important factor for video stabilization.
Any input larger than previous values gives a $50\%$ contribution,
instantly.
The important parameter for this algorithm is the size of $array$, and
the typical value is $8$.
%

There is one last thing that needs to be addressed.
Because equation (\ref{lpf}) accumulates motion data at each
frame, it also accumulates undesirable error motions.
The errors can be derived from motion detection, the approximation of RS
distortion, and the rough decomposition of $N_n^{-1}$.
We should suppress this error accumulation.
We decompose $Q_n$ as in equation (\ref{decomposition}):
\begin{align}
 Q_n \sim R_{\rm yaw}(A_{n}) R_{\rm pitch}(B_{n})
 R_{\rm roll}(\Gamma_{n}) \Lambda_{n},
 \label{Q_decom}
\end{align}
where $A_n,B_n$ and $\Gamma_n$ respectively represent yaw, pitch, and roll for the
distortion-free
cropping matrix.
Only the term $\Lambda_n$ requires special treatment here.
As other terms are
yaw-pitch-roll rotations, they can be treated in the low pass filter with
a minor change.
In the ideal case where every motion is composed of only yaw-pitch-roll
rotations, this $\Lambda_n$
must coincide with the identity matrix.
Hence, to simply suppress the errors, we blend $\Lambda_n$ with
the identity:
\begin{align}
 \Lambda_n \gets (1-\eta)\Lambda_n + \eta I,
\end{align}
%
%
where $\eta$ is a small constant, and the typical value is 0.25.

\section{Stitching stabilizer}
\label{stitching_stabilizer}
\begin{figure}[htbp]
 \centering
 \includegraphics[scale=0.35]{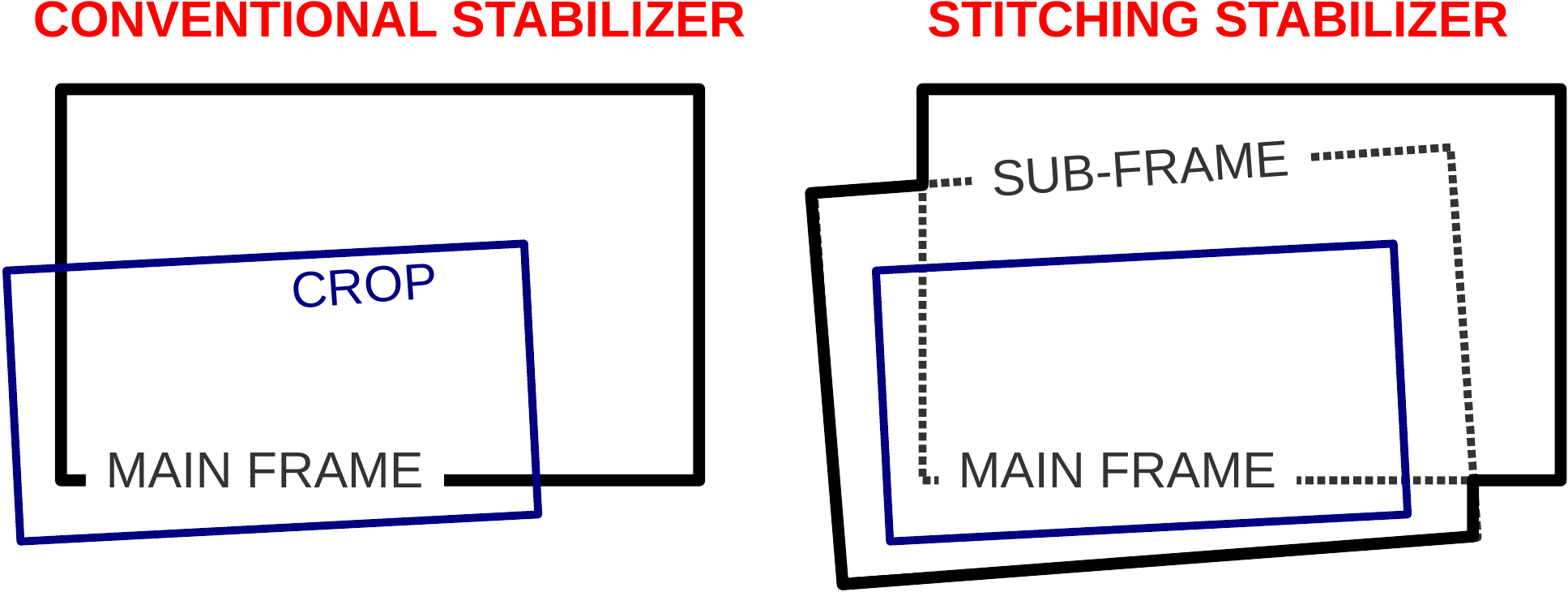}
 \caption{
 In the conventional stabilizer, the cropping region must be inside the
 main frame.
 Hence, there is a deficit region in this cropping configuration, and it cannot be used.
 In the stitching stabilizer, however, this configuration is deficit-free, because
 the cropping region is inside the combined boundary of the main and
 sub-frames.
 Therefore, the stitching stabilizer can absorb camera shake that the
 conventional one cannot.
 }
 \label{inside}
\end{figure}

We explained the algorithm of the conventional
stabilizer.
In this section, we extend it to the stitching stabilizer using
%
the same framework, \alg \ref{realistic_eis}.
%
For simplicity, we rename the current frame as the main frame and
the previous or next frame as the sub-frame.
%
As explained in \fig \ref{inside},
some cropping configurations, that cannot be used in the conventional
stabilizer,
can be used in the stitching stabilizer.
%
%
Hence, we should
replace the \textproc{IsInside} function of \alg \ref{realistic_eis}
with new one.

The new algorithm for \textproc{IsInside} is \alg
\ref{is_inside_stitch}.
The sub-frame boundary is aligned with the main frame to obtain
the combined boundary, and
\textproc{Trans} does this using equation (\ref{drel_n}).
Note that perspective 
transformation makes a rectangle a quadrangle.
\textproc{Combine} combines two quadrangles to return one polygon that
corresponds to the combined boundary.
%
If the cropping boundary is inside the current-previous polygon,
we can get no deficit output with the current-previous stitching.
Therefore, we return true.
The same thing can be said for the current-next case.
%

%
\begin{algorithm}[htbp]
 \caption{\textproc{IsInside} functions for the stitching stabilizer.}
 \label{is_inside_stitch}
 \begin{algorithmic}[1]
  \Function{IsInside}{$crop$}
  \State $M \gets$ \Call{CalcMotion}{from previous to current}
  \If{\Call{IsInsideCombined}{$crop,M$}}
  \State \Return true
  \Comment current-previous stitching
  \EndIf
  \Statex
  \State $M \gets$ \Call{CalcMotion}{from next to current}
  \If{\Call{IsInsideCombined}{$crop,M$}}
  \State \Return true
  \Comment current-next stitching
  \EndIf
  \State \Return false
  \EndFunction
  \Statex
  \Function{IsInsideCombined}{$crop,M$}
  \State $main \gets$ \Call{Rect}{$0,0$,input width,input height}
  \return \Comment the main frame boundary
  \State $sub \gets$ \Call{Trans}{$main,M$}
  \return \Comment the sub-frame boundary aligned with the main
  \State $combined = $ \Call{Combine}{$main,sub$}
  \return \Comment the combined valid boundary
  \If{$crop$ is inside $combined$}
  \State \Return true
  \Else
  \State \Return false
  \EndIf
  \EndFunction
 \end{algorithmic}
\end{algorithm}

Then, let us consider cropping and stitching.
The cropping of the main frame is essentially the same as that of the
conventional stabilizer:
\begin{align}
 (x_n,y_n,1)^{\rm T} \sim P_n (X_n,Y_n,1)^{\rm T}.
\end{align}
Some pixels of $(x_n,y_n,1)$ are outside the main frame, and the
sub-frame
fills in this undefined region.
%
For stitching, the sub-frame has to be aligned with the main frame.
From equation (\ref{drel_n}),
\begin{align}
 (x_{n-1},y_{n-1},1)^{\rm T} \sim M_n^{-1} P_n (X_n,Y_n,1)^{\rm T}, \\
 (x_{n+1},y_{n+1},1)^{\rm T} \sim M_{n+1} P_n (X_n,Y_n,1)^{\rm T}.
\end{align}
These equations transform the output frame coordinates to the sub-frame
coordinates and the alignment can be performed.

As stated, the optimal seam is required for the stitching, and
the algorithm for finding it is explained 
in the next two subsections.
After the optimal seam is found, the merging process is straightforward.
In the stitching stabilizer, \textproc{Crop} in
\alg \ref{realistic_eis} also performs seam-finding and merging, in addition
to cropping.

\subsection{Optimal seam algorithm}
\label{stitching_path}

\begin{figure}[htb]
 \centering
 \includegraphics[scale=0.4]{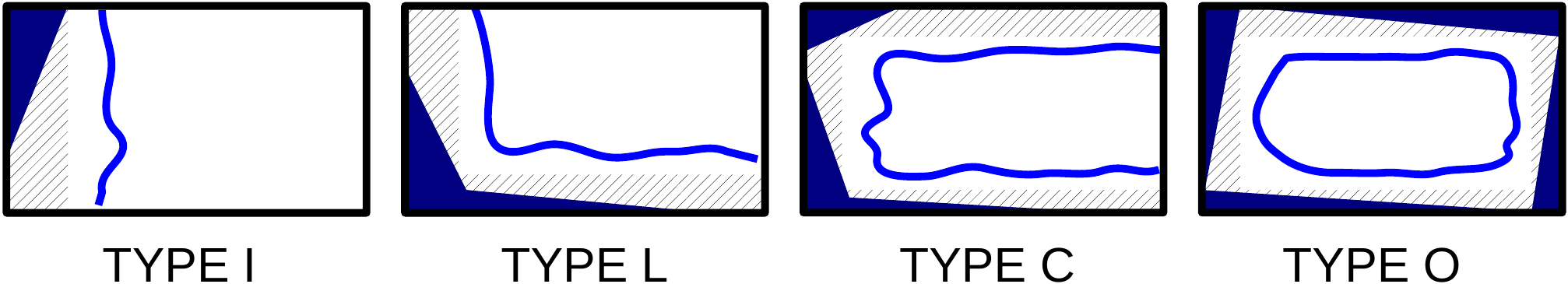}
 \caption{Four types of expanded deficit regions (hatched area),
 examples of the corresponding seam (blue lines), and
 examples of actual deficit regions (blue area).
 We categorize deficit regions to these four types with the minimum expansion.
 Each type I, L, and C includes three variations of its $90^\circ$,
 $180^\circ$, and $270^\circ$ rotations.
 }
 \label{stitching_alg}
\end{figure}

%
For finding the optimal seam,
we categorize each frame with deficit regions into four types, as in
\fig \ref{stitching_alg}.
We expand the deficit region so
that it matches one of these types.
This expansion is performed under the condition that the change of 
the area takes the minimum value.
When deficit regions are not connected,
we virtually connect these regions and assign them to one of these types,
with the minimum expansion.

Here, we compare three algorithms, namely DP, Dijkstra's
algorithm and, graph cuts.
The order of computational complexity of the optimally implemented 
graph cuts
is
$O(VE)$~\cite{orlin2013max}
and that of Dijkstra's algorithm is 
$O(V\log V + E)$~\cite{cormen2009introduction},
where $E$ and $V$ denote the number of edges and nodes, respectively.
The complexity of DP used in this kind of problem is
$O(V)$~\cite{gu2009new}.

The fastest algorithm is desirable when
taking our purpose
into account.
However, DP is one-directional which
means that it can only solve certain cases such as 
those with a top-bottom or left-right seam.
According to our categorization, it can only solve type I and is
therefore not sufficient.
The difference in complexity between DP and graph cuts is huge, so 
graph cuts is a highly undesirable solution.
However, the difference between DP and Dijkstra's algorithm
is not that significant, so they are practically almost the same.
The problem with Dijkstra's algorithm is it can only solve the shortest
path problem and cannot solve type O.
Type O is empirically rare, though, so
we give up type O and use Dijkstra's algorithm for the stitching stabilizer.

Note that we must change the function \textproc{IsInside} in \alg \ref{is_inside_stitch},
because we have given up type O.
The change itself is minor: When
the deficit is type O, we only need to prevent the function from returning true.
The final important factor is that 
we use $1/4$ shrunken images to accelerate the processing time.

\subsection{Finding the optimal seam}
\begin{figure}[htbp]
 \centering
 \includegraphics[scale=0.3]{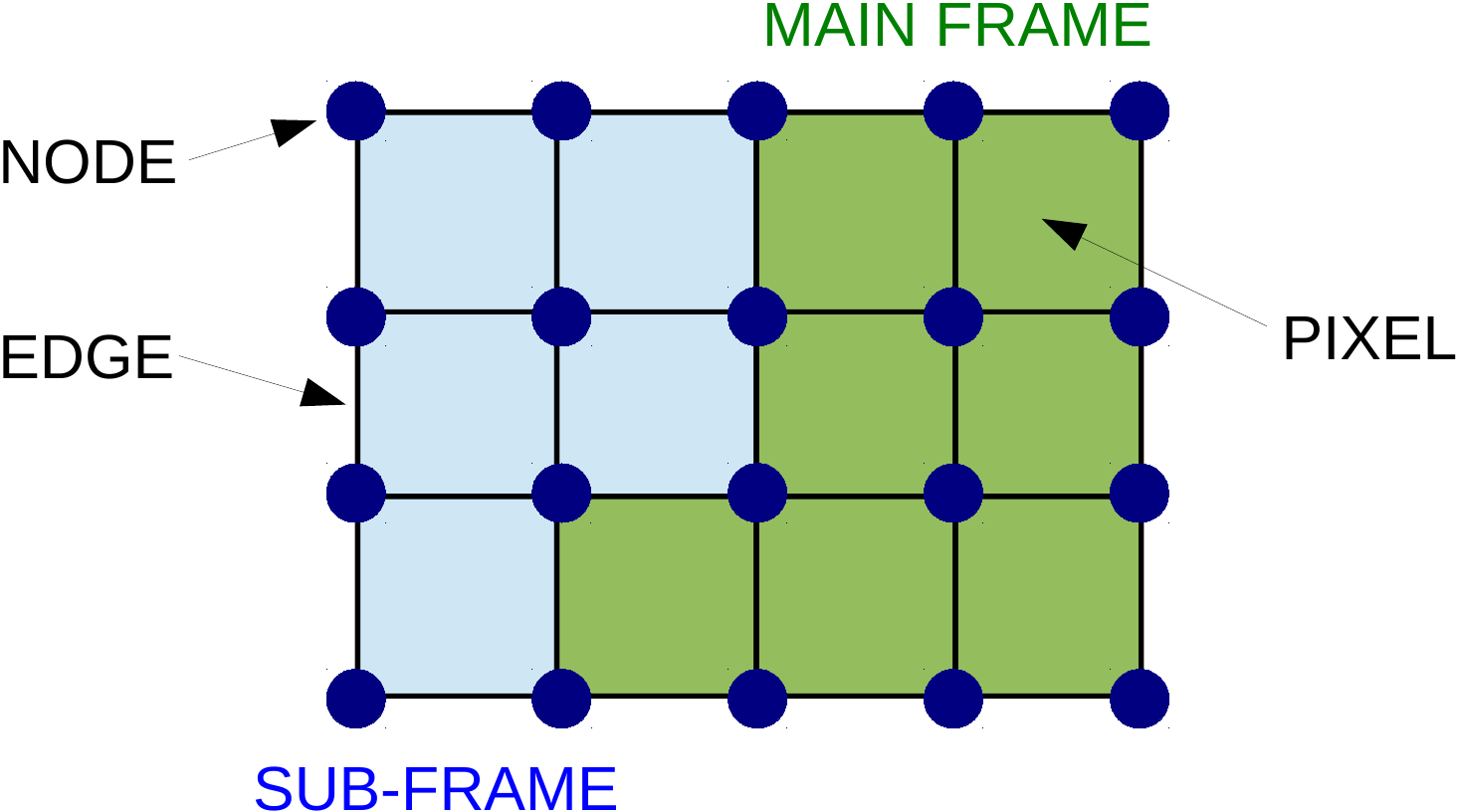}
 \caption{
 In our search with Dijkstra's algorithm,
 nodes are the criss-crossed
 points of pixels and edges are the border lines of pixels.
 Costs are only assigned to edges.
 }
 \label{cost}
\end{figure}

To find the seam with Dijkstra's algorithm,
we must define what the nodes, 
edges, and costs mean in our case.
We explain this using \fig \ref{cost}.
The edges are defined as the borders of image pixels and the nodes are
the criss-crossed points of four neighboring pixels.
We assign costs only for edges:
%
%
\begin{align}
 {\rm cost}_{AB}
 &=
 \left| L_A^{\rm main} - L_B^{\rm sub} \right| +
 \left| L_A^{\rm sub} - L_B^{\rm main} \right| ,
 \label{vcost}
\end{align}
where $L_A$ and $L_B$ denote the luma values for any pair of adjacent pixels, and
$cost_{AB}$ is the cost of the edge between them.
With this definition, the cost becomes small when there is only a small difference in the
luma values across the seam.
Hence, the resulting seam might be the most unnoticeable one
for stitching.
The cost is set to infinity in the vicinity of the deficit
region so that the seam does not cross it.
On the border of the output frame, the edge cost cannot be determined by the
above equations. We set these border costs to zero, except for those inside
the deficit region.

\begin{figure}[htbp]
 \centering
 \includegraphics[scale=0.4]{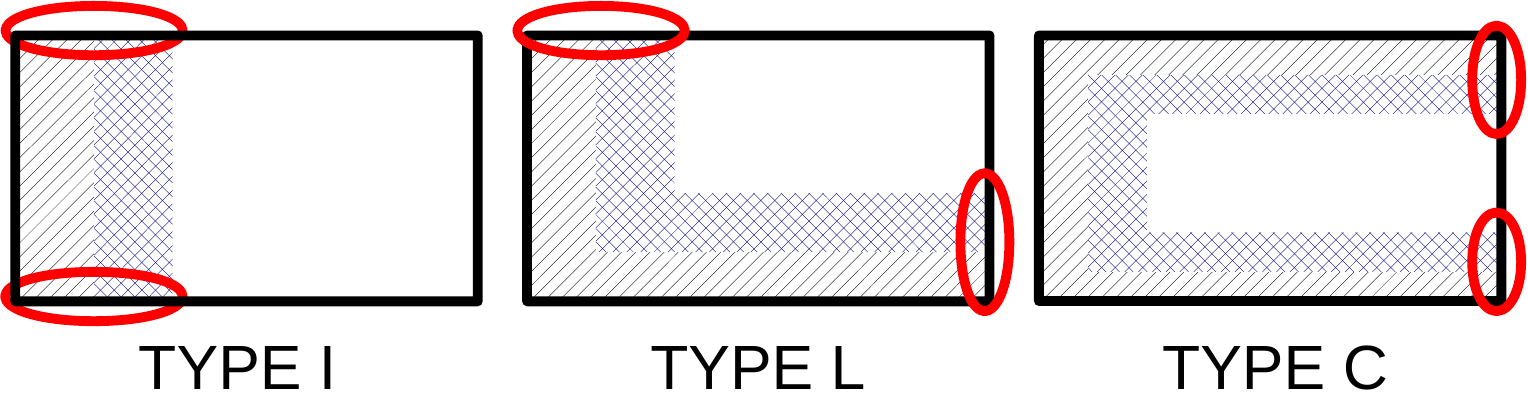}
 \caption{Expanded deficit region (sparsely hatched region), 
 search region (both densely blue hatched and sparsely hatched regions), and start and end points
 (circled in red).
 We set the search region using the type of the expanded
 deficit region.
 In our implementation, the width of the search region was set $2-4$ times wider than that of
 the expanded deficit region.}
 \label{start_end}
\end{figure}

Then, we set the search region and the start and end points.
Using the main frame is generally preferable to using the sub-frame.
Hence, the optimal seam should be drawn near the perimeter of
the main frame.
The simplest way to enforce this is to restrict the search region, which
is also beneficial for the processing time.
As depicted in \fig \ref{start_end}, we set the search region based on
the shape and size of the expanded deficit region.
Both ends of the
search region are assigned to start or end points.

After setting the costs, the search region and the start and end points,
we can apply  Dijkstra's algorithm to find the optimal seam for stitching.

\section{Hyperlapse}
\label{hyperlapse}
In this section, the extension of the stitching stabilizer to hyperlapse
videos is explained.
To process hyperlapse in mobile devices, it is possible to perform
a preprocess during the video capture.
In our case, global motion detection is done as the preprocess.
This is because, it is
highly advantageous for achieving strong stabilization, to know all global motions of the clip in
the main process.
%
During the video capture, the function \textproc{CalcMotion} in \alg
\ref{realistic_eis} is called every frame, 
and its result, the motion matrix $M$, is
stored as a kind of meta data.
%

\alg \ref{hyperlapse_process} is the main hyperlapse process.
This is similar to the standard stabilization algorithm,
\alg \ref{realistic_eis}.
Here, {\it standard} means the algorithm explained in sections
\ref{motion_detection_and_filtering} and \ref{stitching_stabilizer}.
The difference is only in the loop of the image frame and motion data
skipping. Here, $skip$ is the number of skipped frames.
In the stitching hyperlapse, the sub-frame corresponds to the
$skip$-frame previous or next frame.
\textproc{LoadMotion} loads the corresponding
global motion data that were stored by the preprocess algorithm.
%
\begin{algorithm}[htbp]
 \caption{Main process for hyperlapse.}
 \label{hyperlapse_process}
 \begin{algorithmic}[1]
  \State $P \gets identity$
  \While{there is a frame to be processed}
  \State $M \gets identity$
  \For{$i \gets 1, skip$}
  \Comment frame skipping
  \State Load the next input frame
  \State $M \gets M \times$ \Call{LoadMotion}{from previous to current}
  \EndFor
  \State $P \gets$ \Call{Filter}{$M, P$}
  \State $P \gets$ \Call{EnsureInside}{$P$}
  \State \Call{Crop}{$P$}
  \EndWhile
 \end{algorithmic}
\end{algorithm}

\subsection{Filter design for hyperlapse}
\label{stabilization_for_hyperlapse}
Part of the low pass filter should be redesigned so that it
matches hyperlapse videos.
Specifically, we change filter $g$ in equation
(\ref{filter_g}).
The most important difference between standard stabilization and
hyperlapse is that we can utilize all global motion data including
future data.

We call filtered rotational values in equation
(\ref{filter_g}), namely $g(\alpha_n),g(\beta_n)$, and $g(\gamma_n)$, 
{\it camera velocities}.
We consider that the ideal output for hyperlapse is an output with constant
camera velocities.
This is difficult to achieve in many realistic cases, however, so we have to accept
several {\it turns}.
In addition, the stabilizer should crop the near-center regions
of the input frames, if possible.
This is because, the object that users want to capture might be in the center of the frame.
Needless to say, a highly undesirable output is what goes outside the
input frame, and \textproc{EnsureInside} works to make sure
it is inside.
In consideration of the conditions above, we try to minimize the following cost
to get the filtered values of $\alpha_n,\beta_n$, and
$\gamma_n$ for each frame:
\begin{align}
 {\rm cost} &= 
 \varepsilon_t \times \text{number of turns}\nonumber \\
 &+
 \sum_{i \in \text{all frames}} \left( 
 \varepsilon_\alpha {\rm abs}(A_i) +
 \varepsilon_\beta {\rm abs}(B_i) +
 \varepsilon_\gamma{\rm abs}(\Gamma_i) 
 \right)  \nonumber \\
 &+
 \varepsilon_p \times \text{number of attempts} \nonumber \\
 &\qquad \qquad
 \text{to crop the outside area of the frames},
\end{align}
where $A_i,B_i$, and $\Gamma_i$ are defined in 
equation (\ref{Q_decom}).
The first term increases the cost when turns occur; the second term
represents the cost for cropping the off-center region; and the third term adds
penalties for trying to crop the outside the frames.
The typical values of the parameters are 
$\varepsilon_t =1,\varepsilon_p = 16,\varepsilon_\alpha=\varepsilon_\beta=\varepsilon_\gamma=1/10^\circ$, 
although it depends on the camera parameters.

%
%
We search the possible choices for camera velocities to get the result with
the minimum cost.
However, one problem is that 
the number of possible choices is infinite.
Therefore, we perform a simplified and finite possibility version of the
search, \alg \ref{finite_search}.
In this algorithm, we use a tree, $tree$.
For every frame, we add new nodes to the leaves of $tree$ and each added
node represents how the stabilizer works at that frame.
Hence, each path from the root to a leaf represents a sequence of
stabilization results.
The node contains its parent node, virtual stabilizer, camera
velocities, and cost.

Before the process,
we append some initial nodes to the root.
In every frame, we only consider the leaves with the maximum depth, that
correspond to the nodes of the previous frame.
This is because, all current frame nodes must be appended to the
previous frame nodes.
For each maximum-depth leaf, two possibilities are considered. The first is
continuing the constant motion with the stored velocities~(line
\ref{constant_motion}), and the second is
making turns with new velocities~(line \ref{new_turn}).
The new velocities are determined by averaging input motion $\alpha_n,\beta_n$,
and $\gamma_n$ with different periods.
$T$-times turns are attempted for any leaf, and
\textproc{AverageMotion}$(p)$ returns the averaged motions with the period $p$.
\textproc{Process}$(stabilizer,veloc)$ performs the stabilization
process explained in
sections \ref{motion_detection_and_filtering} and \ref{stitching_stabilizer}, for the virtual stabilizer
$stabilizer$ with the camera velocities $veloc$.
\textproc{CalcCost} returns the off-center cost and the outside penalty.
The number of new nodes are limited to $S$, because, without this
limitation, 
the number increases exponentially.
The stored nodes are determined by the costs.
Then, the new nodes are appended to the leaves of $tree$.
After all frames are processed, we select the leaf with
the minimum cost, and use the corresponding path for the desired camera velocities.

\begin{algorithm}[htb]
 \caption{Finite search algorithm for the hyperlapse filter.}
 \label{finite_search}
 \begin{algorithmic}[1]
  \State $tree \gets$ new Tree
  \State Add some initial nodes to the root
  \While{there is a frame to be processed}
  \State \Call{SearchEachFrame}{$tree$}
  \State Load the data of the next frame
  \EndWhile
  \State Get the path with the minimum cost
  \Statex

  \Function{SearchEachFrame}{$tree$}
  \State $d \gets 2$
  \State $period[] \gets \{ d^1, d^2, d^3, d^4, d^5,...\}$
  \State $leaves \gets $ Get leaves with the maximum depth
  \State $list \gets$ new List
  \ForAll{$node \in leaves$}
  \State $veloc \gets node\text{.velocities}$
  \State $list$.add(\Call{NewNode}{$node,veloc,0$})
  \label{constant_motion}
  \return \Comment constant motion

  \For{$t \gets 1, T$}
  \State $veloc \gets$ \Call{AverageMotion}{$period[t]$}
  \State $list$.add(\Call{NewNode}{$node,veloc,\varepsilon_t$})
  \label{new_turn}
  \EndFor
  \Comment try new turns
  \EndFor

  \State \Call{Sort}{$list$}
  \Comment sort by costs
  \For{$s \gets 1, S$}
  \State $tree$.add($list$.get($s$))
  \EndFor
  \Comment add only the best $S$ nodes to the leaves
  \EndFunction
  \Statex

  \Function{NewNode}{$node,veloc,\varepsilon$}
  \State $stabilizer \gets node\text{.stabilizer}$
  \State \Call{Process}{$stabilizer,veloc$}
  \State $cost \gets node\text{.cost} + \varepsilon +$ \Call{CalcCost}{$stabilizer$}
  \State \Return new Node($node,stabilizer,veloc,cost$)
  \return \Comment (parent, branched stabilizer, camera velocities, cost)
  \EndFunction
 \end{algorithmic}
\end{algorithm}

%
%
This limited search needs to process all image frames to obtain
the first result, and
it is unsuitable for long video clips.
Hence, we put more restrictions on the search and
separate the process for the first frame from that
for others frames.
This is shown in \alg \ref{first_search}.
%
%
For the first frame result, we select the minimum cost leaf $best\_n$ immediately after the $N$-th
frame is processed, and determine the path from the root to $best\_n$.
%
We use the node next to the root, on this path, 
as the first frame result.
In this way, the first result can be obtained by processing only $N$
frames.
We replace $tree$ by its subtree with this result node as the new root,
because all future nodes must be descendants of this result.

For other frames, we call \textproc{SearchEachFrame} only once,
not $N$-times as for the first frame,
and get the result by \textproc{FixState}.
In other words,
we fix the result at $m$ by finding the best leaf after the $(m+N)$-th process,
in a frame-by-frame manner.
Note that the height of $tree$ is always kept to $N$ by creating
subtrees.
The computational cost for the first frame is much
higher than that for other frames.
In other words, our algorithm requires additional processing time for
creating the first output and this is studied in subsection \ref{processing_time}.
As parameters, we typically set $N=64,S=1024$, and $T=6$.
\begin{algorithm}[htb]
 \caption{Finite search algorithm, revised.}
 \label{first_search}
 \begin{algorithmic}[1]
  \State $tree \gets$ new Tree
  \State Add some initial nodes to the root
  \State $ret \gets$ \Call{GetFirst}{$tree$}
  \Comment this is used for the first result
  \While{there is a frame to be processed}
  \State $ret \gets$ \Call{GetOthers}{$tree$}
  \Comment process frame-by-frame
  \EndWhile
  \Statex

  \Function{GetFirst}{$tree$}
  \For{$i \gets 1, N$}
  \State \Call{SearchEachFrame}{$tree$}
  \State Load the data of the next frame
  \EndFor
  \State \Return \Call{FixState}{$tree$}
  \EndFunction
  \Statex

  \Function{GetOthers}{$tree$}
  \State \Call{SearchEachFrame}{$tree$}
  \Comment process $(m+N)$-th data
  \State Load the data of the next frame
  \State \Return \Call{FixState}{$tree$}
  \Comment get $m$-th result
  \EndFunction
  \Statex

  \Function{FixState}{$tree$}
  \State $best\_n \gets$ Get the best leaf at the maximum depth
  \State $path\_n \gets$ Get the path from the root to $best\_n$
  \State $ret \gets $ Get the node next to the root on $path\_n$
  \State $tree \gets $ Make the subtree of $tree$ with $ret$ as the new root
  \return \Comment keep only the nodes which are descendants of $ret$
  \State \Return $ret$
  \EndFunction
 \end{algorithmic}
\end{algorithm}

%

\section{Result and discussion}
\label{results}
In this section, we show several results of our stitching stabilizer and discuss them.
%
%
We used Full HD~(1920x1080) and 30 FPS videos for both the input and
output in all of the results described.

\subsection{Stitching example}
\label{stitching_example}
We show some examples of stitching process here.
We used the results of $\times 4$ hyperlapse in order to
show the differences clearly.

\begin{figure*}[htbp]
 \centering
 \includegraphics[scale=0.21]{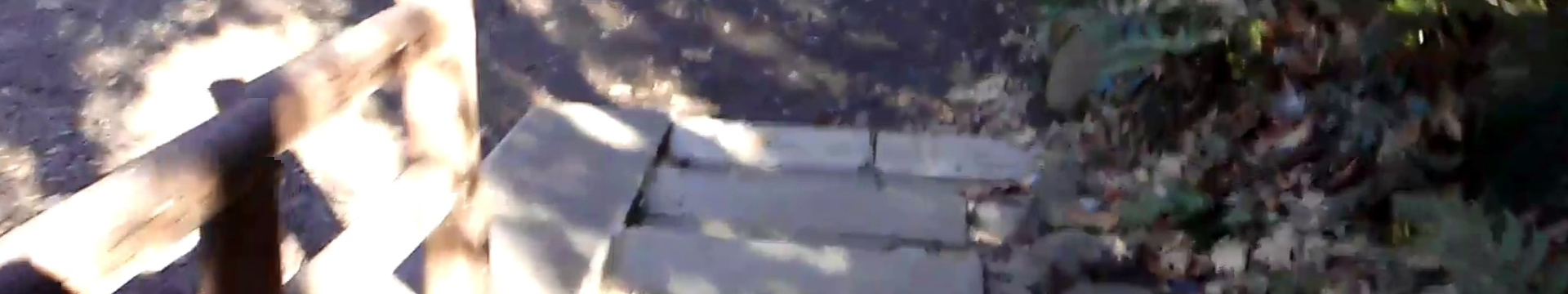}
 \includegraphics[scale=0.21]{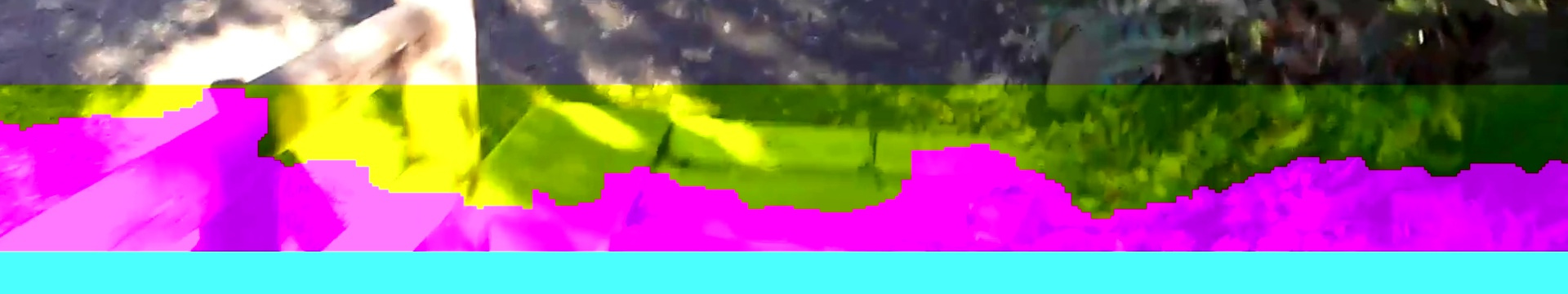}
 \includegraphics[scale=0.21]{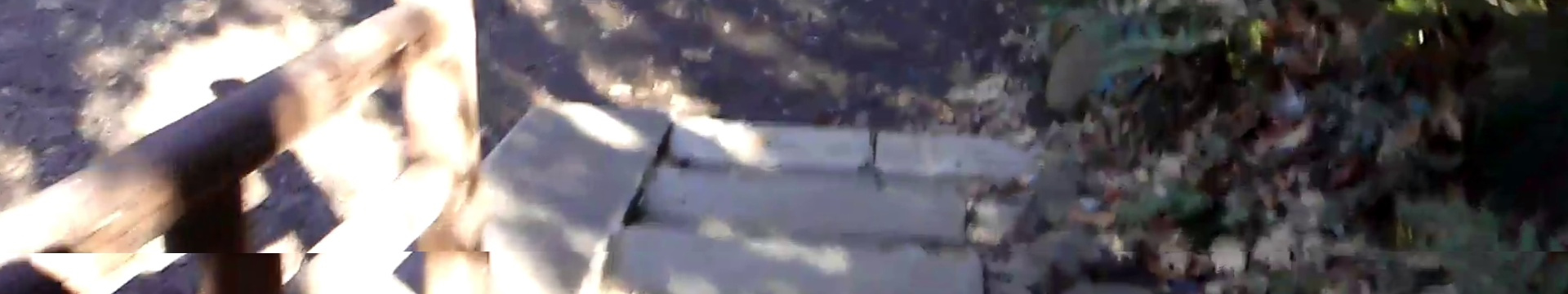}
 \caption{
 The top panel is the stitching result;
 the middle panel shows the search region (yellow), the
 region pasted from the sub-image (pink) and the deficit region (blue);
 the bottom panel is the result of naive stitching where the deficit
 region is simply filled by the sub-image.
 The boundary of the yellow and pink regions corresponds to the
 optimal seam.
 While an artificial line is visible in the naive result, it is much
 less noticeable
 in the result of our stitcher.}
 \label{stitching_example1}
\end{figure*}

\begin{figure*}[htbp]
 \centering
 \includegraphics[scale=0.21]{02output_crop.jpg}
 \includegraphics[scale=0.21]{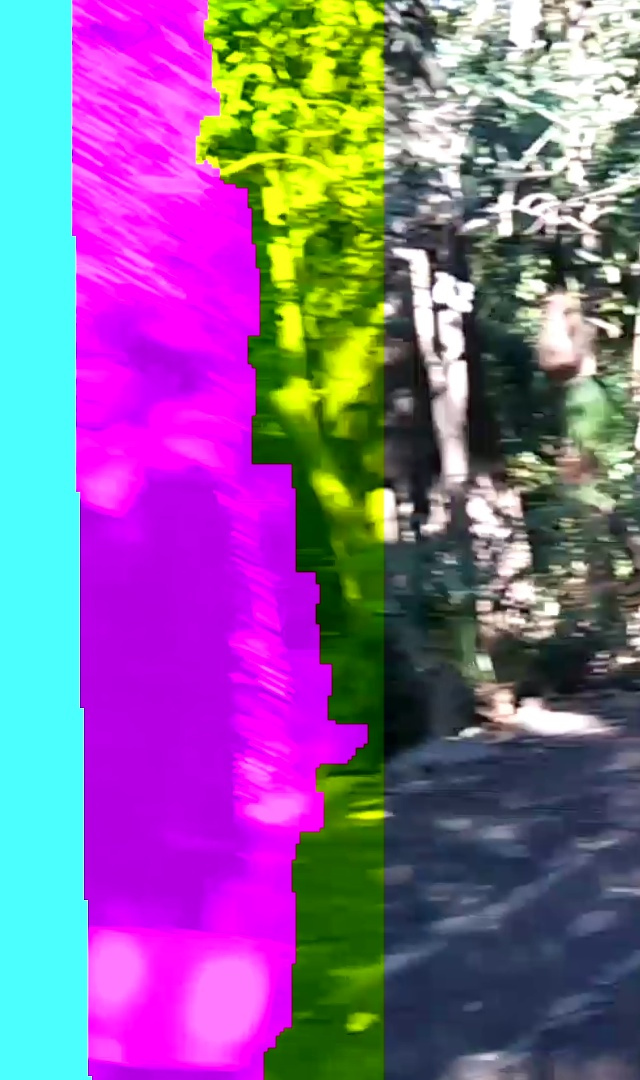}
 \includegraphics[scale=0.21]{02naive_crop.jpg}
 \caption{
 The left panel is the stitching result;
 the center panel shows the search region (yellow), the
 region pasted from the sub-image (pink) and the deficit region (blue);
 the right panel is the result of naive stitching where the deficit
 region is simply filled by the sub-image.
 Our stitcher prevents the human face from being cut 
 in half.}
 \label{stitching_example2}
\end{figure*}

Three images were prepared for
both \fig \ref{stitching_example1} and \ref{stitching_example2}:
the result of our stitcher,
the image showing
the deficit region and the optimal seam, and
the result of naive stitching where the deficit
region is simply filled by the sub-image.
These figures show that our stitcher clearly reduce the unnaturalness of the naive result.
It is especially effective for non-planar parts of the scene such as a foreground
object or a person walking.

\subsection{Standard stitching stabilizer}
\label{real-time_video_stabilization}

\begin{figure}[htbp]
 \centering
 \begin{subfigure}[t]{\seqsize \linewidth}
  \centering
  \includegraphics[scale=0.05]{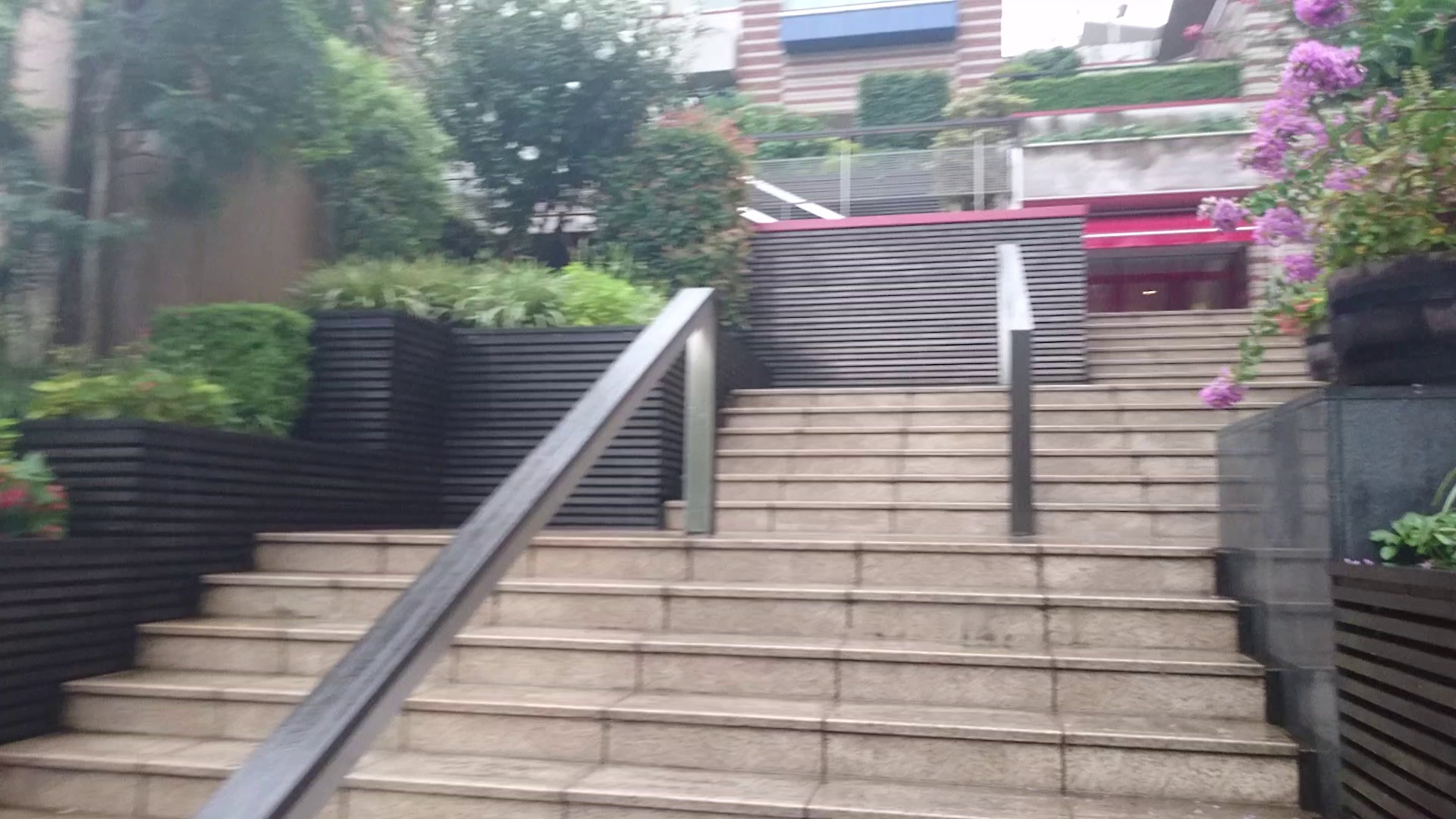}
  \caption*{
  \centering
  Sequence 1: Walking (upstairs)

  by Xperia~Z3.}
 \end{subfigure}
 \begin{subfigure}[t]{\seqsize \linewidth}
  \centering
  \includegraphics[scale=0.05]{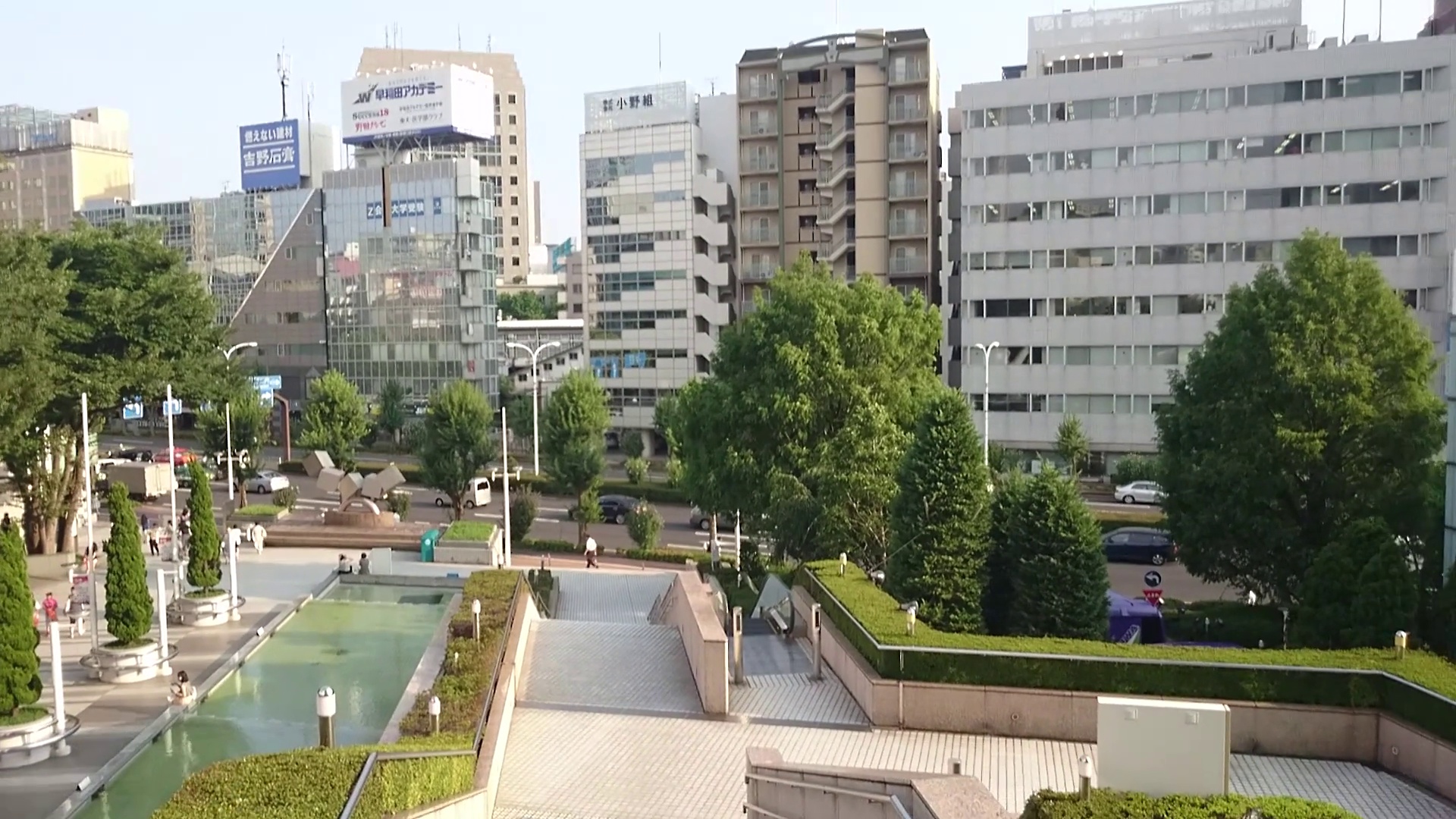}
  \caption*{
  \centering
  Sequence 2: Walking (downstairs)

  by Xperia~Z3.}
 \end{subfigure}
 \begin{subfigure}[t]{\seqsize \linewidth}
  \centering
  \includegraphics[scale=0.05]{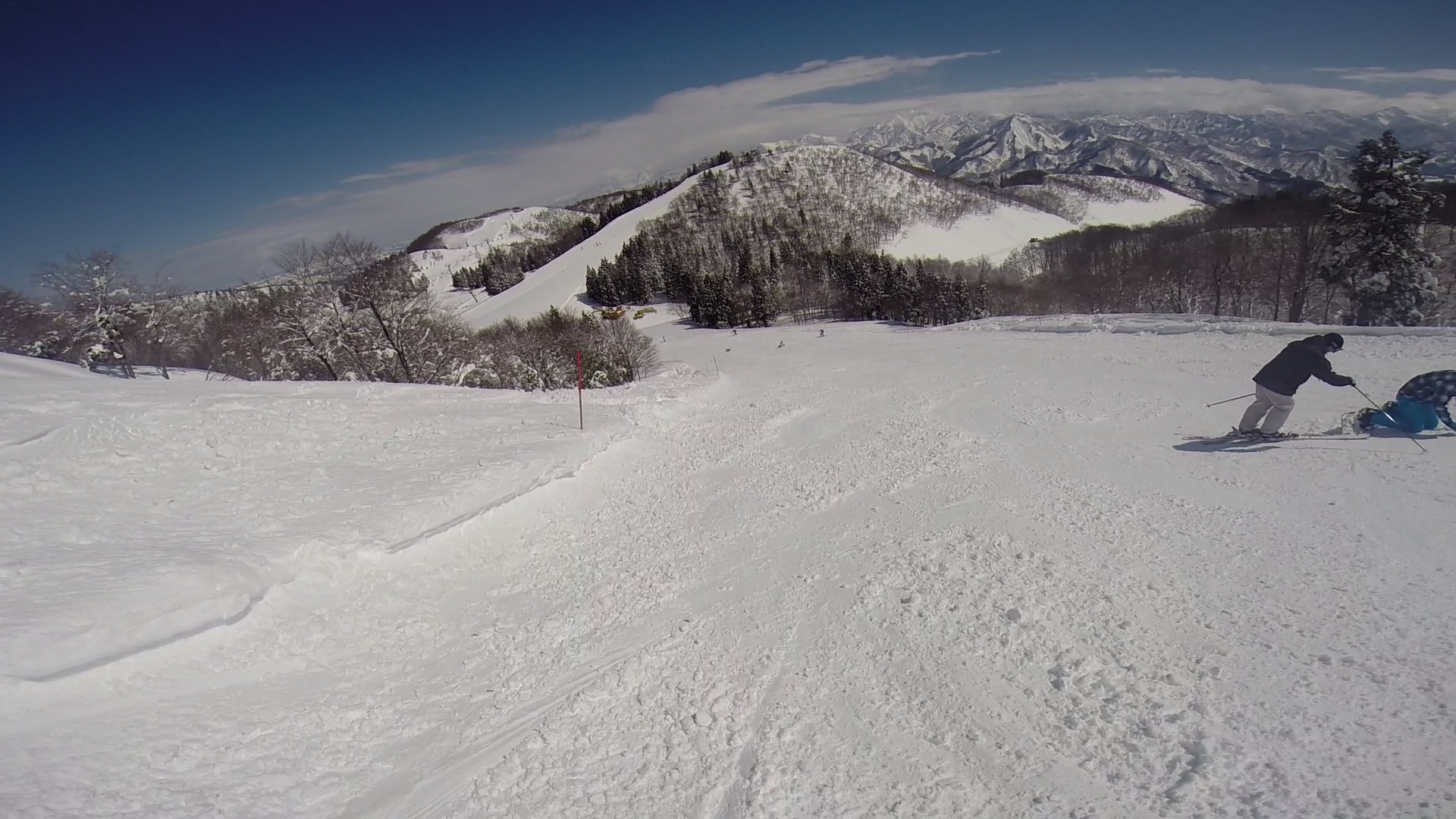}
  \caption*{
  \centering
  Sequence 3: Skiing

  by GoPro~HERO3.}
 \end{subfigure}
 \caption{Sample sequences. These are
 typical inputs for video stabilization.}
 \label{real-time_seq}
\end{figure}

We briefly evaluated the standard stitching stabilizer
using three sequences, as shown in \fig
\ref {real-time_seq}.
These videos were taken by devices without using any stabilization.
Because our objective is to achieve a wide AOV, 
we set the cropping ratio to $90\%$, which corresponds 
to 1728x972 pixels.
The standard choice is $80\%$.

We quantify the improvement in the stitching stabilizer by using the number of
frames $n_f$ where \textproc{EnsureInside} works.
Note that stabilization is imperfect in such frames, because this
function introduces artificial shake to the result.
We show this in \tab \ref{real_nf}.
%
The stitching stabilizer reduces $n_f$, especially in sequences 1 and 2.
Therefore, it works properly.
\begin{table}[htbp]
 \centering
 \caption{Difference in the number of failed frames $n_f$ between
 the conventional and stitching stabilizer.
 }
 \label{real_nf}
 \begin{tabular}{|c|c|c|c|}
  \hline
  && \multicolumn{2}{|c|}{$n_f$} \\
  \cline{3-4}
  & All frames & Conventional & Stitching  \\
  \hline
  Sequence 1 & 1550 & 510 & 176 \\
  Sequence 2 & 1196 & 456 & 139 \\
  Sequence 3 & 1608 & 623 & 322 \\
  \hline
 \end{tabular}
\end{table}

%



\subsection{Hyperlapse}
\label{results_hyperlapse}

We also set the cropping ratio to 90\%, and used the sequences in \fig \ref{real-time_seq} to evaluate
the stitching stabilizer in hyperlapse.
%
%
Because the filtering algorithm tries to avoid the case where \textproc{EnsureInside}
works with full effort,
it is difficult to
quantify the effect of the stitching stabilizer in the same manner as in
the standard case.
Instead, we used the final cost value for comparison,
since desirable algorithms might choose lower cost results.
Our searching algorithm does not
guarantee that the cost of the stitching stabilizer is smaller than that
of the conventional one, so this should be tested.
The result is in \tab \ref{hyper_cost},
where we used two parameter sets for filtering.
The standard set is $N=64,S=1024,T=6$ and the lite set
is $N=64,S=256,T=6$.
The cost of the standard set is smaller than that of the
lite set,
and the cost of the stitching stabilizer is smaller than that of the
conventional one, as expected.
%
\begin{table*}[htbp]
 \centering
 \caption{Difference in the final costs between the conventional and
 stitching stabilizers.}
 \label{hyper_cost}
 \begin{tabular}{|c|c|c|c|c||c|c|c|c|}
  \hline
  & \multicolumn{4}{|c||}{$\times 4$ Hyperlapse}
  & \multicolumn{4}{|c|}{$\times 8$ Hyperlapse} \\
  \cline{2-9}
  & \multicolumn{2}{|c|}{Standard}
  & \multicolumn{2}{|c||}{Lite}
  & \multicolumn{2}{|c|}{Standard}
  & \multicolumn{2}{|c|}{Lite} \\
  \cline{2-9}
  & Conv. & Stit. 
  & Conv. & Stit.
  & Conv. & Stit. 
  & Conv. & Stit.  \\
  \hline
  Sequence 1
  & 1932.896 & 514.436
  & 2156.461 & 592.443
  & 2377.071 & 980.790
  & 2420.753 & 999.795 \\
  Sequence 2
  & 2262.367 & 258.584
  & 2388.724 &  260.431
  & 1566.641 &  144.600
  & 1728.087 &  169.681 \\
  Sequence 3 
  & 2512.996 & 991.503 
  & 2796.446 & 1190.797
  & 2525.599 & 1402.273
  & 2539.572 & 1590.828 \\
  \hline
 \end{tabular}
\end{table*}

In the supplemental video\footnote{\protect\url{https://www.youtube.com/watch?v=LmyPXfGZRb0}}, we included an exerpt of the result of
sequence 2.
We prepared the input video with naive frame skipping, the result of
the conventional stabilizer, and that of the stitching stabilizer,
for $\times 4$ hyperlapse using the standard parameter set.
%
In this video, it is clear that the stitching stabilizer reduces camera shake more effectively
than the conventional one does, in hyperlapse.

\begin{figure}[htbp]
 \centering
 \includegraphics[scale=0.1]{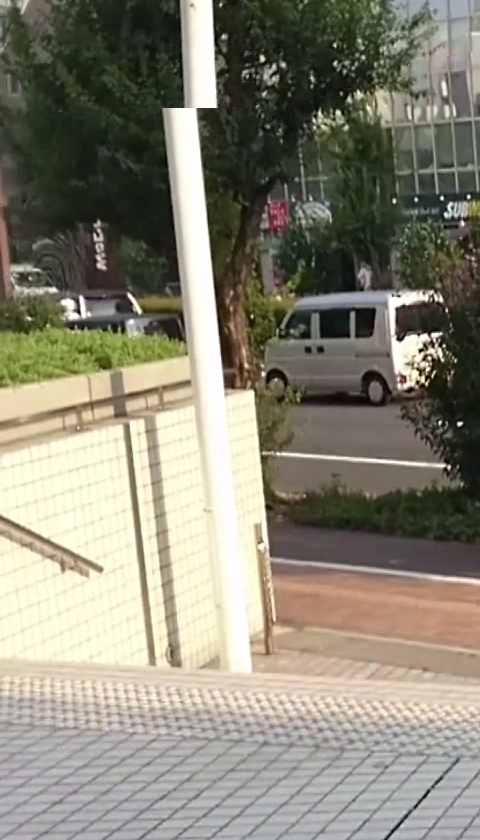}
 \caption{A pole is cut in half in an output frame of sequence 2.
 The stitching stabilizer does not work perfectly for large or long foreground objects.}
 \label{limitation}
\end{figure}

In the result of sequence 2, there is a frame where a pole is cut in half,
and we show a part of the frame in \fig \ref{limitation}.
It is difficult to stitch perfectly for
large or long foreground objects, such as a pole in this example, 
This is because, the optimal seam cannot circumvent such large objects.
Although this is a limitation of the stitching stabilizer,
we think it is not a crucial problem for hyperlapse.
The reason is following.
First, hyperlapse is a kind of fun features.
Second, even if this problem occurs in several frames, it is not an easily 
noticeable defect in first-forwarding videos.
For the standard stitching stabilizer, this issue is rare, because the temporal
distance between the main and sub-frame is relatively small.
In that case, non-planar behavior of foreground objects is
manageable in the stitching stabilizer algorithm.

\begin{figure}[htbp]
 \centering
 \begin{subfigure}[t]{\seqsize \linewidth}
  \centering
  \includegraphics[scale=0.05]{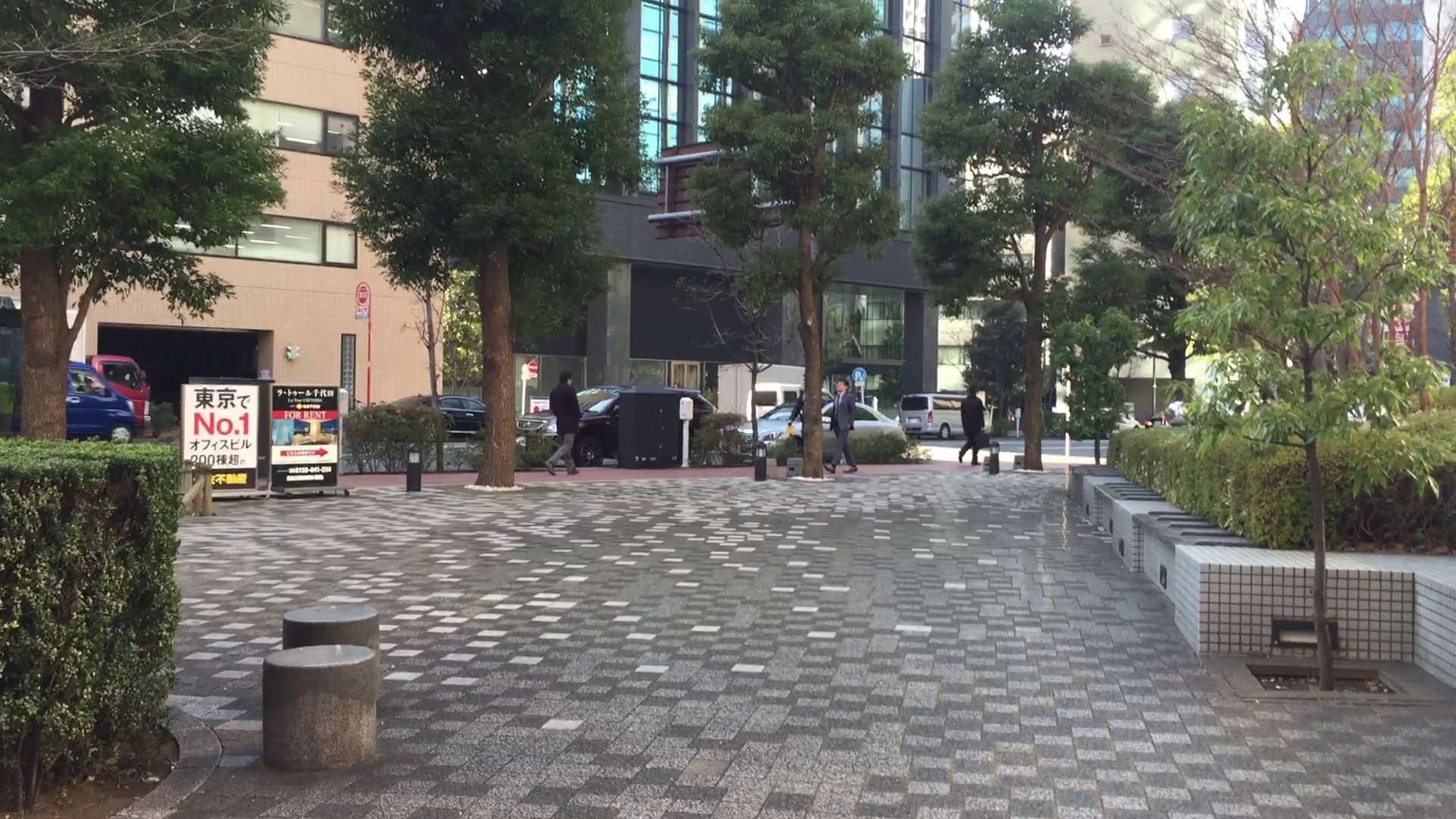}
  \caption*{
  \centering
  Sequence 4: Jogging~($\times 4$)

  by iPhone~6.}
 \end{subfigure}
 \begin{subfigure}[t]{\seqsize \linewidth}
  \centering
  \includegraphics[scale=0.05]{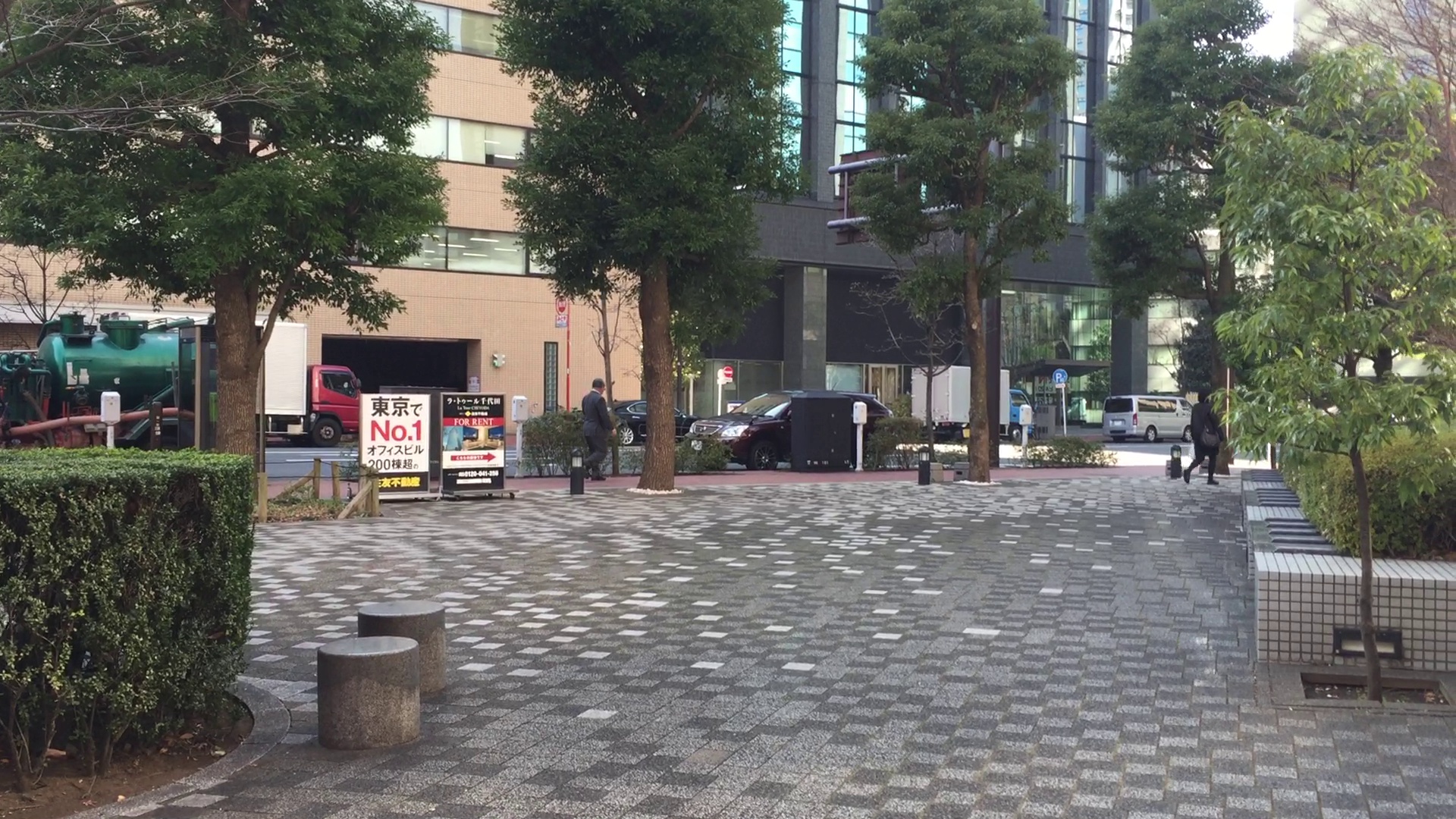}
  \caption*{
  \centering
  Sequence 5: Jogging~($\times 8$)

  by iPhone~6.}
 \end{subfigure}
 \begin{subfigure}[t]{\seqsize \linewidth}
  \centering
  \includegraphics[scale=0.05]{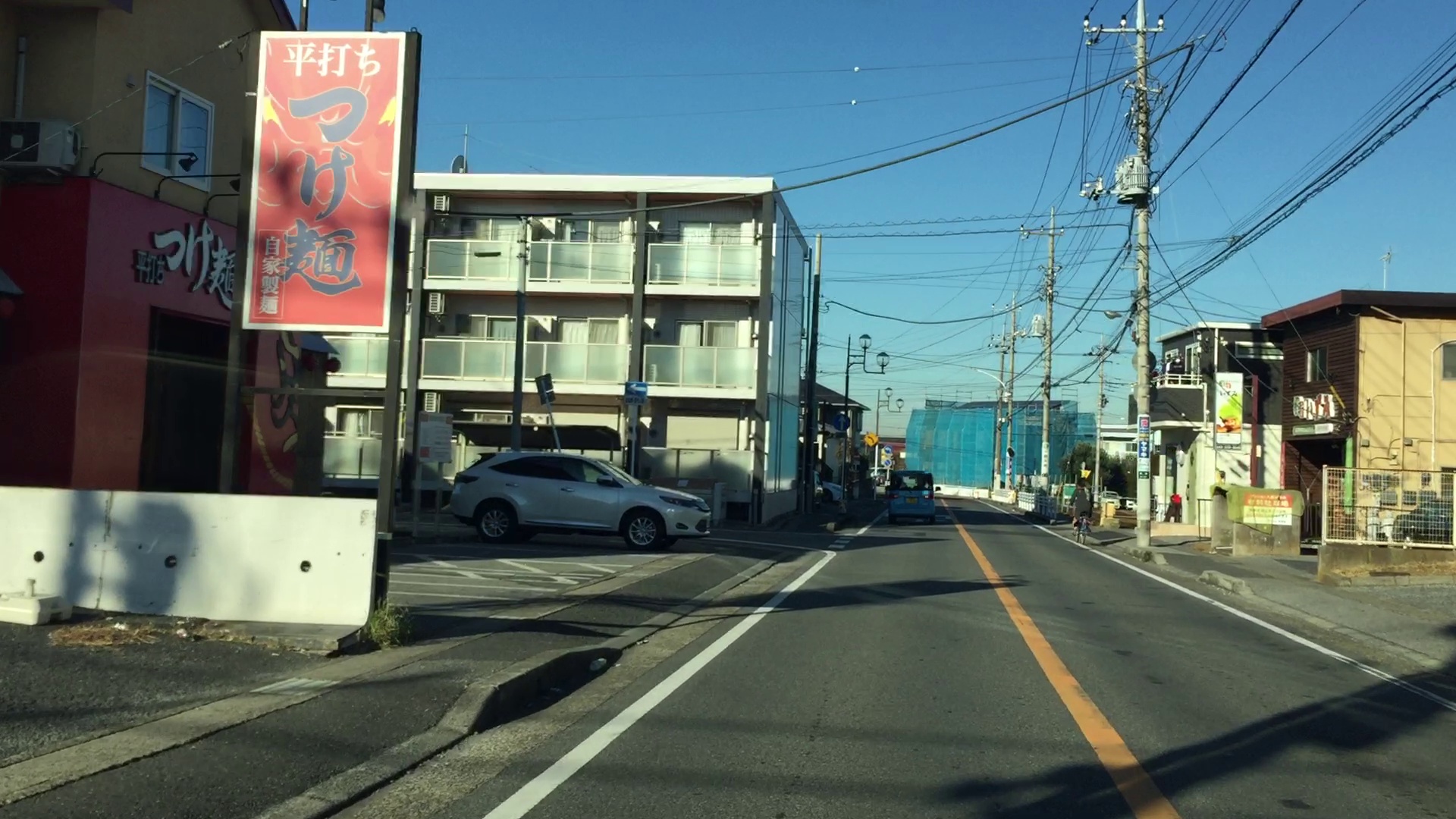}
  \caption*{
  \centering
  Sequence 6: Driving~($\times 4$)

  by iPhone~6.}
 \end{subfigure}
 \caption{Sequences used for comparison with apps of Microsoft and Instagram.
 The clips of sequence 4 and 5 are taken at the same day and on the same route.
 The difference is the setting of hyperlapse, $\times 4$ and $\times 8$.}
 \label{comp_seq}
\end{figure}

For hyperlapse, real-time apps are available for comparison:
Microsoft's mobile app~(Microsoft
Hyperlapse Mobile)~\cite{joshi2015real}
and
Instagram's app~(Hyperlapse from
Instagram)~\cite{karpenko2014introduction}.
Although it is not a real-time solution, 
Microsoft's PC app~(Microsoft Hyperlapse Pro)~\cite{kopf2014first}
might also be a comparison target.
We compared the stitching stabilizer with these existing solutions
by using several sequences in \fig \ref{comp_seq}.
For the comparison, we use the cropping ratio of 80\%, which corresponds to
1536x864, with the standard parameter set.
We used two iPhone~6 devices, which were fixed to the same frame, to take videos.
This is because Instagram's app does not accept external files, and 
we cannot retrieve the input video from the app.
Hence, we used the first device only for Instagram's app.
The results for the stitching stabilizer and Microsoft's apps were
created from the same video file taken using the second device.
The stitching stabilizer and Instagram's app perform
simple frame skipping, but Microsoft's apps do not.
Therefore, we cannot maintain synchronization between the results in comparison videos.
%
%
%
%

We created
$\times 4$ hyperlapse videos for sequences 4 and 6, and $\times 8$ for
sequence 5, and compared the results.
An exceprt of the most apparent result, that of sequence 4, 
is included in the supplemental video.
Microsoft's apps, including the PC app, show wobbling and unnatural acceleration
and deceleration.
Although Instagram's result is natural, its stabilization is weak.
The stitching stabilizer produces a more stabilized result than the mobile
apps of Instagram and Microsoft,
and the result is more natural than that achieved using Microsoft's apps.
Moreover, at least for this sequence, the result of the stitching stabilizer
is even more desirable than that of Microsoft's PC app, a non
real-time solution.
The results of the other sequences show essentially the same charasteristic.
Hence, we can say that the stitching stabilizer
produces a more stabilized and natural output than the existing
solutions.

Note that more results of the stitching stabilizer can be found in 
the supplemental video, where we used 80\% cropping with the standard
parameter set.


\subsection{Processing time}
\label{processing_time}
%
We measured the processing time of the stitching stabilizer using
a commercial device, since it is a crucial issue for real-time solutions.
We used Sony's Xperia Z2 Tablet, which was first launched in the spring
of 2014 and has a Qualcomm MSM8974 chipset~(Snapdragon 801), as the device.
We ran the executable on this device without rooting.
%
%
Because this was not a new device,
if the stitching 
stabilizer runs in realistic time on this device, it could be used in
a variety of recent handsets.
We only utilized a CPU with vectorization and
parallel threading and did not perform any special tuning
for this specific model.

For the standard stitching stabilizer, we measured the first 256 frames for each
sequence in \fig \ref{real-time_seq}, and calculated the average FPS.
The results are listed in \tab \ref{time_conventional}.
%
The values of FPS here are sufficient enough for 30 FPS video stabilization.

\begin{table}[htbp]
 \centering
 \caption{Average FPS
 of the standard stitching stabilizer.}
 \label{time_conventional}
 \begin{tabular}{|c|c|}
  \hline
  & \shortstack{Process~[FPS]} \\
  \hline
  Sequence 1
  & 64.758 \\
  Sequence 2
  & 54.124 \\
  Sequence 3
  & 49.978 \\
  \hline
 \end{tabular}
\end{table}

We also measured the time for hyperlapse with the same sequences.
The results are summarized in \tab \ref{time_hyperlapse}.
The first 256 frames were used for measurement.
When there were fewer than 256 output frames,
we used all of the frames for the measurement of the main process.
The first latency means the time required to create the first
output, as
explained in subsection \ref{stabilization_for_hyperlapse}.
The values of preprocessing FPS are much larger than 30 and 
fast enough to be
run during the video capture.
Although the FPS values for main process are large, the first
latency also takes a large value.
This means we need to wait about half a second to start the
hyperlapse with the standard parameter set and about 200 msec with the
lite set.
There is no explicit threshold for the latency,
and it depends on the situation.
Hence, the latency should be taken into account when tuning the parameters.
%
\begin{table*}[htbp]
 \centering
 \caption{Average FPS of hyperlapse with standard and lite parameter sets.}
 \label{time_hyperlapse}
 \begin{tabular}{|c|c||c|c|c|c||c|c|c|c|}
  \hline
  &
  & \multicolumn{4}{|c||}{Main process~[FPS]}
  & \multicolumn{4}{|c|}{First latency~[msec]} \\
  \cline{3-10}
  &
  & \multicolumn{2}{|c|}{$\times 4$ Hyperlapse}
  & \multicolumn{2}{|c||}{$\times 8$ Hyperlapse}
  & \multicolumn{2}{|c|}{$\times 4$ Hyperlapse}
  & \multicolumn{2}{|c|}{$\times 8$ Hyperlapse} \\
  \cline{3-10}
  & \shortstack{Preprocess~[FPS]}
  & \shortstack{Standard}
  & \shortstack{Lite}
  & \shortstack{Standard}
  & \shortstack{Lite}
  & \shortstack{Standard}
  & \shortstack{Lite}
  & \shortstack{Standard}
  & \shortstack{Lite} \\
  \hline
  Seq. 1
  & 68.129
  & 76.805
  & 100.210
  & 72.056
  & 107.712
  & 423.615
  & 173.182
  & 752.164
  & 236.283\\
  Seq. 2
  & 69.109
  & 89.350
  & 130.310
  & 64.425
  & 89.270
  & 527.586
  & 245.783
  & 451.672
  & 240.983\\
  Seq. 3
  & 70.602
  & 98.135
  & 100.664
  & 74.716
  & 56.142
  & 551.502
  & 163.183
  & 644.948
  & 239.036 \\
  \hline
 \end{tabular}
\end{table*}

The stitching stabilizer has not yet been fully optimized,
and the results listed above was measured on a device that was not new.
Therefore, it could be executed with much faster processing time on
recent smartphones.




\section{Conclusion}
\label{conclusion}

We introduced the {\it stitching stabilizer}, a software-based video stabilization
algorithm that stitches two adjacent input
frames together.
The stitching process effectively expands the area of input frames and achieves more powerful
stabilization than the conventional single-frame algorithm.
While the existing methods~\cite{litvin2003probabilistic,matsushita2006full,chen2008capturing}
also do this,   
their computational cost is too high for embedded systems.
The main focus of our algorithm is a real-time process
for embedded systems such as smartphones.
%
%

%
We applied the stitching stabilizer to hyperlapse.
Because of fast-forwarding, camrera shake is increased in hyperlapse videos.
%
The result of the stitching stabilizer was much smoother than that of the
non-stitching stabilizer.
We also showed that the stitching stabilizer created a more strongly stabilized and
natural output than the frame selection algorithm~\cite{joshi2015real}
and Instagram's algorithm~\cite{karpenko2014introduction}, which also focus on
real-time processing.
The results of the stitching stabilizer was even more desirable than that of 
Kopf et al.~\shortcite{kopf2014first},
in some cases.

By using a commercial device, we showed that the processing time 
of the stitching stabilizer is fast enough for recent smartphones.

The current algorithm
works well in many cases. However, there is a fundamental
difficulty in some cases, such as when there are large or long foreground objects.
In such cases, the simple two-frame stitching algorithm is not very effective.
It seems practical to implement a checker that
decides whether the scene is suitable for stitching.

Stitching the current frame and the
two-frame previous or two-frame next frame, instead of one, might
provide stronger stabilization,
because we can get a larger input frames by fusing temporally
distant frames.
While this might introduce unnatural effects, it is worth trying.
Stitching more than two frames is another possibility, yet with more
computational costs.

For hyperlapse, 
the stitching stabilizer is further enforced with a frame selection
algorithm such as one by Joshi et al.
This is a possible direction of research for achieving a higher quality
hyperlapse videos.
%